\documentclass[11pt]{article}

\usepackage[final]{acl}

\usepackage{times}
\usepackage{latexsym}

\usepackage[T1]{fontenc}

\usepackage[utf8]{inputenc}
\usepackage{CJKutf8}
\usepackage{microtype}

\usepackage{inconsolata}

\usepackage{graphicx}
\usepackage{tcolorbox}
\usepackage{enumitem}
\usepackage{geometry}
\usepackage{amssymb}
\usepackage{mathrsfs}
\usepackage{amsmath}
\usepackage{hhline}
\usepackage{amsfonts}
\usepackage{multirow}
\usepackage{tabularx}
\usepackage{array}
\usepackage{booktabs}
\usepackage{verbdef}
\usepackage{colortbl}
\usepackage{color}
\usepackage{subcaption}
\usepackage{threeparttable}
\usepackage{float}
\usepackage{caption}
\usepackage{arydshln}
\usepackage{rotating}
\usepackage{xcolor}
\usepackage[normalem]{ulem}
\usepackage{placeins}
\usepackage{makecell}
\usepackage{afterpage}
\usepackage{blindtext}
\usepackage{pifont}
\usepackage{threeparttablex}
\usepackage[framemethod=TikZ]{mdframed}
\usepackage{hyperref}
\usepackage{marvosym}

\title{Summarization Is Not Dead Yet}
\author{
 \textbf{Dongqi Liu\textsuperscript{$\Omega$}\textsuperscript{$\Theta$}\thanks{\scalebox{1.3}{\Letter} dongqi.me@gmail.com}},
 \textbf{Chenxi Whitehouse\textsuperscript{$\Delta$}},
 \textbf{Zheng Zhao\textsuperscript{$\Gamma$}},
  \\
 \textbf{Zhuchen Cao\textsuperscript{$\Omega$}},
 \textbf{Jian Li\textsuperscript{$\Theta$}},
 \textbf{Yabiao Wang\textsuperscript{$\Psi$$\Theta$}}
 \\
 \textsuperscript{$\Omega$}Saarland University, Max Planck Institute for Informatics,
 \textsuperscript{$\Delta$}University of Cambridge \\
 \textsuperscript{$\Gamma$}University of Edinburgh, 
 \textsuperscript{$\Psi$}Zhejiang University,
 \textsuperscript{$\Theta$}Tencent YouTu Research \\
}
\begin{document}
\maketitle

\begin{abstract}
The progress of large language models (LLMs) has fueled claims that model-generated summaries rival or even surpass human-written references, raising questions about whether summarization remains an open research problem. We re-examine this narrative through a multi-track evaluation covering diverse datasets and state-of-the-art LLMs, combining controlled human assessment, bias-mitigated LLM-as-Judge protocols, factuality verification against external knowledge, and corpus-level linguistic analysis. Our findings reveal a more nuanced landscape in which human references continue to demonstrate advantages in informativeness and faithfulness, whereas LLM outputs are preferred mainly for surface-level coherence and fluency. Factuality verification indicates that human references remain more reliable, particularly for claims involving reasoning or synthesis, and linguistic analysis uncovers a pattern of stylistic homogeneity across different models. These observations suggest that current LLMs have raised the floor of summarization quality, but the ceiling of their performance remains below human capabilities.
\end{abstract}

\section{Introduction}
Summarization has long been a popular research area in natural language processing (NLP), concerned with condensing source inputs (e.g., long documents and video recordings) into shorter representations that preserve salient information \cite{10.1145/3731445, pu-demberg-2024-rst}. With the advent of LLMs, the landscape of summarization has shifted considerably \cite{liu-etal-2024-learning}, as these models deliver strong summarization capabilities \cite{zhang-etal-2024-benchmarking, fonseca-cohen-2024-large-language, ravaut-etal-2024-context}. Several studies have also reported that LLM-generated summaries tend to be preferred over human-written summaries and may achieve comparable or superior factual consistency \cite{liu-etal-2024-learning, liu-etal-2023-revisiting, goyal2023newssummarizationevaluationera, pu2023summarizationalmostdead}. These statements have raised the question of whether \textit{\uline{continued research in summarization remains warranted}} and whether \textit{\uline{core challenges of summarization have been largely addressed by general-purpose LLMs}}.

\begin{figure}[t]
  \centering
  \includegraphics[width=\linewidth]{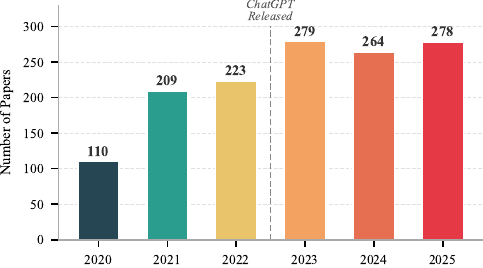}
  \caption{Number of summarization papers at major NLP venues (2020–2025). Papers are identified by the presence of \textit{summarization} or \textit{summarisation} in the title or abstract; each paper is counted once.}
  \label{fig:summ_papers_by_year}
\end{figure}

Yet, publication trends at major NLP venues point to the fact that summarization research remains active. As shown in \autoref{fig:summ_papers_by_year}, the number of summarization-related papers at TACL, ACL, EMNLP, NAACL, EACL, and AACL has grown from 110 in 2020 to 278 in 2025, with expanding directions including controllable generation \cite{he-etal-2022-ctrlsum, pu-demberg-2023-chatgpt, urlana-etal-2024-controllable}, multimodal investigation \cite{liang-etal-2023-summary, papalampidi-lapata-2023-hierarchical3d, liu-etal-2025-talk}, and trustworthy evaluation \cite{chrysostomou-etal-2024-investigating, wan-etal-2024-acueval, cao-etal-2022-hallucinated, pu-etal-2023-incorporating}. The view that LLMs have largely solved summarization nonetheless rests on three recurring observations: LLM-generated summaries (i) receive higher overall preference rates in human evaluation, (ii) outperform human summaries under LLM-as-Judge protocols, and (iii) contain fewer hallucinations than human summaries \cite{pu2023summarizationalmostdead, liu-etal-2024-learning, zhang-etal-2024-benchmarking}.

Several methodological factors limit the generalizability of these conclusions: (i) Holistic preference judgments that ask annotators to pick a preferred summary without separating quality dimensions may conflate surface fluency with information density, favoring LLM outputs even when coverage is insufficient \cite{liu-etal-2023-g, min-etal-2025-towards}. (ii) LLM-as-Judge evaluations may introduce position bias and self-preference bias, which undermine reliability if not carefully controlled \cite{zheng2023judging, li-etal-2025-generation}. (iii) Source-only factuality protocols penalize extrinsic content as hallucination regardless of whether it reflects valid world knowledge or unfaithful content. Since human-written summaries more often add background content as intentional contextualization while LLM outputs more often add it through generation-time fabrication, this conflation systematically disadvantages human answers \cite{cao-etal-2022-hallucinated, liu-etal-2025-explanatory}. In this work, we employ dimension-specific evaluation with larger annotator pools, apply bias-mitigated LLM-as-Judge protocols, verify factuality against external knowledge sources, conduct linguistic analysis at lexical, syntactic, and discourse levels, and extend evaluations across multimodal, multilingual, and style-constrained settings.

\textbf{Our main contributions are as follows:}
\begin{itemize}[leftmargin=8pt,itemsep=1pt,topsep=1pt,parsep=1pt]
\item We conduct controlled human and bias-mitigated LLM-as-Judge assessments across multiple domains and show that human summaries retain advantages in informativeness and faithfulness when assessed independently of surface fluency.
\item We introduce factuality verification against external knowledge to separate legitimate world knowledge from genuine hallucinations and reveal that LLM summaries still contain a proportion of errors, especially for claims that require reasoning or synthesis beyond the source text.
\item We perform a comprehensive linguistic analysis at lexical, syntactic, and discourse levels and find that LLM summaries exhibit lower lexical diversity, shallower syntactic structures, and weaker discourse-level organization.
\end{itemize}
Collectively, our findings support the central thesis that \textit{summarization is not dead yet}. While LLMs have raised the floor of summarization quality, the human ceiling has not yet been reached. By surfacing these persistent gaps, we provide a grounded basis for the community to identify open problems and prioritize future directions in summarization research.

\section{Related Work}
We organize prior work into two strands, one arguing that advances in LLMs reduce the need for summarization research, and another documenting persistent limitations that challenge this view. As discussed below, each strand examines a limited set of quality dimensions within a narrow range of domains, and neither offers a controlled, multi-dimensional comparison across diverse settings that would more fully inform this discussion. This study aims to contribute toward addressing this limitation.

\paragraph{Arguments That Summarization Is Solved.}
The most direct articulation of this position is offered by \citet{pu2023summarizationalmostdead}, who conduct a human preference study comparing zero-shot LLM outputs with human-written summaries. Their results indicate a preference for LLM-generated summaries, leading the authors to conclude that conventional summarization research is no longer necessary in the era of LLMs. In a related study, \citet{zhang-etal-2024-benchmarking} evaluate LLMs on news summarization and find that instruction tuning plays an influential role in zero-shot performance. When fresh references written by professional freelancers are introduced, LLM summaries are judged to be broadly comparable to human-written ones. \citet{adams-etal-2023-sparse} propose \textit{Chain of Density} prompting, in which GPT-4 iteratively rewrites summaries to increase entity density without extending length, and find that annotators prefer intermediate-density CoD summaries whose entity density is close to that of human-written references. \citet{yao-etal-2023-improving} frame practical summarization research as shifting toward post-editing and human–AI collaboration rather than fully autonomous generation. \citet{wang-etal-2025-empirical} show that zero-shot LLMs achieve competitive performance against fine-tuned traditional models on many-to-many summarization.

\paragraph{Arguments That Summarization Remains Open.}
A body of work challenges the ``solved'' narrative across several interconnected dimensions. \citet{panickssery2024llm} uncover that LLMs exhibit systematic self-preference bias, inflating scores for outputs resembling their own generation. Broader analyses identify structural biases, including familiarity bias and score-anchoring effects, which tend to reward fluent and generic text \citep{wang-etal-2024-large-language-models-fair, wan-etal-2025-positional}. Evaluating long-form summarization reliably remains methodologically unsettled \citep{guo-vosoughi-2023-length, kim2024fables, chang2024booookscore, belem-etal-2025-single}. Multilingual settings introduce additional challenges, with standard metrics showing reduced reliability for non-English languages \citep{qiu-etal-2023-detecting,forde-etal-2024-evaluating}, and standard annotation procedures underestimating error rates \citep{min-etal-2025-towards}. In clinical settings, LLMs lack agreed-upon standards for acceptable summaries \citep{croxford2025current, nagar-etal-2025-umedsum}. In code summarization, LLM attention patterns diverge from those of human programmers \citep{li2024machines}. At a more fundamental level, human summaries consistently integrate deeper reasoning and inferential compression, whereas LLM outputs operate largely through sophisticated paraphrasing \citep{zeweniuk-etal-2025-beyond}.

\section{Evaluation Setup}
\label{sec:setup}

To examine our claims, we design a multi-track evaluation that compares human reference summaries with outputs from five state-of-the-art LLMs, namely GPT-5.4 (GPT), Claude Opus 4.6 (Claude), Gemini 3.1 Pro Preview (Gemini), Qwen3.5-397B-A17B (Qwen), and Kimi-K2.5 (Kimi). The evaluation spans five datasets (CNNSum, SciNews, DiverseSumm, VISTA, and EurLexSum), encompassing ultra-long documents, style transfer settings, multi-document inputs, multimodal content, and multilingual scenarios. 

We generate one summary per instance for each closed-source LLM (GPT, Claude, and Gemini) via the provider's API. For open-source models (Qwen and Kimi), we use the HuggingFace Inference API under the same evaluation protocol. Decoding is greedy ($T = 0$), with a maximum of 1{,}024 new tokens per completion; top-$p$, top-$k$, and frequency penalties retain each API's defaults. Each dataset uses a domain- and language-specific instruction template (full prompts in \autoref{appendix:prompts}); within a dataset, all five models receive the identical prompt, so performance differences reflect model capability rather than prompt variation. 

\autoref{tab:dataset_samples} reports the per-track sample counts. Human evaluation uses a random subsample to manage annotation cost, while automatic evaluations use all available samples. Detailed information on model identifiers and dataset descriptions is provided in \autoref{appendix:setup_details}. The evaluation protocols are documented in \autoref{appendix:human_eval} through \autoref{appendix:linguistic_setup}.

\begin{table}[h]
\centering
\small
\resizebox{0.95\linewidth}{!}{%
\begin{tabular}{lcc}
\toprule
Dataset & Human Eval & Automatic Eval \\
\midrule
CNNSum & 300 & 695 \\
SciNews & 300 & 4,188 \\
DiverseSumm & 200 & 245 \\
VISTA & 300 & 1,859 \\
EurLexSum (per lang.) & 30 & 188 \\
EurLexSum (total) & 720 & 4,512 \\
\midrule
Total & 1,820 & 11,499 \\
\bottomrule
\end{tabular}
}
\caption{Sample counts by dataset and evaluation track. The automatic tracks comprise LLM-as-Judge evaluation, factuality verification, and linguistic analysis.}
\label{tab:dataset_samples}
\vspace{-10pt}
\end{table}

\section{Do Human Evaluators Prefer LLMs?}
\label{sec:human_eval}

We recruit human annotators from the Prolific platform to rate summaries on four 1-to-5 Likert dimensions (informativeness, faithfulness, coherence, conciseness) and to rank the six candidates (one human reference and five model outputs) by overall quality. All candidates are presented in a blind and randomized order; each sample is independently assessed by three crowd annotators, and inter-annotator agreement is monitored via Krippendorff's $\alpha$ ($\geq 0.7$). Full annotation guidelines, quality control procedures, and derivation details are provided in \autoref{appendix:human_eval_setup}.

\autoref{fig:human_pairwise} presents pairwise win-rate matrices for each dataset. Human win rates against the five models range from 0.66 to 0.90.\footnote{Pairwise win rates are derived from listwise overall rankings, with the derivation procedure detailed in \autoref{appendix:human_eval_setup}.} \autoref{fig:human_agreement} breaks these gaps down by dimension. Human summaries hold positive gaps on informativeness and faithfulness ($+$0.17 to $+$0.51) but negative gaps on coherence against GPT and Claude ($-$0.13 to $-$0.25), indicating that annotators perceive model summaries as more fluent on average.

\begin{figure*}[t]
  \centering
  \includegraphics[width=\textwidth]{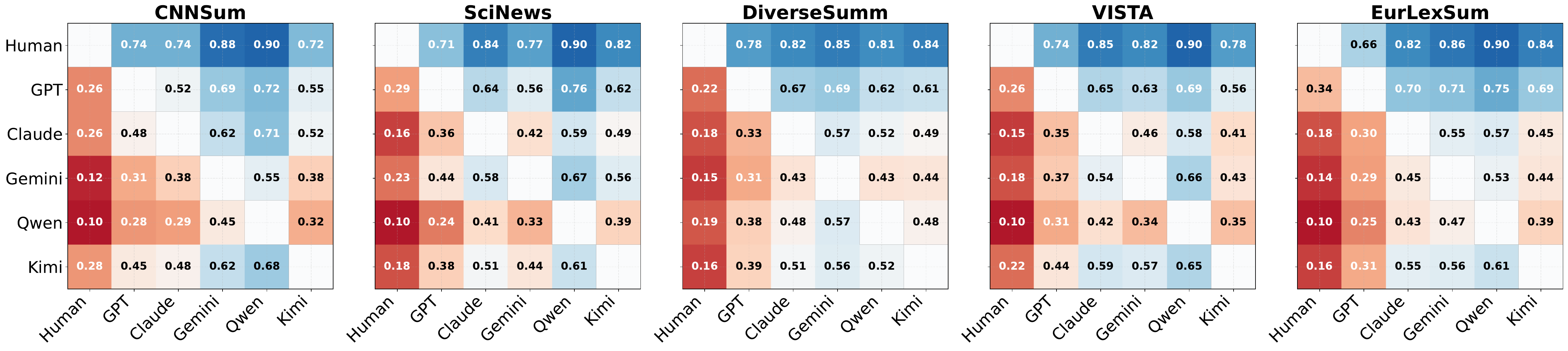}
  \caption{Pairwise win rates from human evaluation, shown separately for each dataset. Each cell reports the proportion of samples for which crowd annotators rank the row system above the column system, averaged across annotators within each sample.}
  \label{fig:human_pairwise}
\end{figure*}

\begin{figure*}[t]
  \centering
  \includegraphics[width=\textwidth]{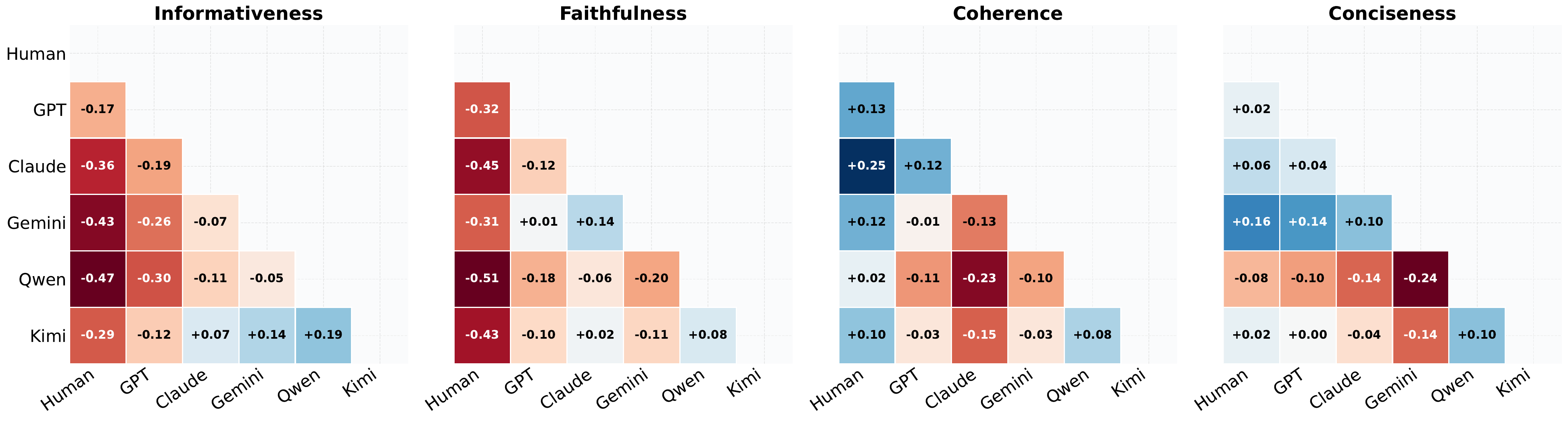}
  \caption{Likert score differences from human evaluation across four dimensions. Each lower-triangular cell reports the mean score of the row system minus that of the column system, averaged across datasets after first averaging annotators within each sample.}
  \label{fig:human_agreement}
\end{figure*}

This dimension-dependent behavior helps explain why prior studies relying on holistic preference judgments often conclude that LLM summaries are competitive with human-authored summaries. Surface fluency and information density are hard to disentangle within a single overall assessment, and dimension-specific evaluation surfaces trade-offs that holistic judgments leave implicit \cite{min-etal-2025-towards, song-etal-2024-finesure}. These findings are robust to generation choices, with prompt specificity and decoding temperature (\autoref{appendix:prompt_sensitivity}) and chain-of-thought or self-correction prompting (\autoref{appendix:prompt_strategies}) keeping human win rates on informativeness and faithfulness.

\paragraph{Contamination-Free Human Summaries.} We again recruit crowd workers under the same qualification criteria (\autoref{appendix:human_eval_setup}), and present them with the same task prompt used to generate model summaries (\autoref{appendix:prompts}). Annotators may consult credible external resources such as Wikipedia, academic papers, and books to support comprehension of the source material, but are explicitly prohibited from using any AI tool. Each worker is given a 120-minute window per sample. We collect 24 controlled human summaries per dataset (120 in total); for EurLexSum, the 24 samples cover all 24 official EU languages. All these controlled summaries are evaluated under the same Likert and listwise protocol.

\begin{table*}[h]
\centering
\small
\setlength{\tabcolsep}{25pt}
\resizebox{\textwidth}{!}{%
\begin{tabular}{llcccc}
\toprule
Dataset & Condition & Info.\ $\Delta$ & Faith.\ $\Delta$ & Coher.\ $\Delta$ & Conci.\ $\Delta$ \\
\midrule
\multirow{2}{*}{CNNSum}
  & Original reference  & $+$0.40 & $+$0.34 & $-$0.10 & $-$0.06 \\
  & Contamination-free   & $+$0.45 & $+$0.38 & $-$0.08 & $-$0.03 \\
\midrule
\multirow{2}{*}{SciNews}
  & Original reference  & $+$0.34 & $+$0.37 & $-$0.11 & $-$0.09 \\
  & Contamination-free    & $+$0.43 & $+$0.45 & $-$0.08 & $-$0.05 \\
\midrule
\multirow{2}{*}{DiverseSumm}
  & Original reference  & $+$0.38 & $+$0.32 & $-$0.06 & $+$0.07 \\
  & Contamination-free    & $+$0.44 & $+$0.38 & $-$0.08 & $+$0.08 \\
\midrule
\multirow{2}{*}{VISTA}
  & Original reference  & $+$0.35 & $+$0.29 & $-$0.12 & $-$0.05 \\
  & Contamination-free   & $+$0.39 & $+$0.32 & $-$0.09 & $-$0.03 \\
\midrule
\multirow{2}{*}{EurLexSum}
  & Original reference  & $+$0.45 & $+$0.38 & $-$0.05 & $-$0.10 \\
  & Contamination-free    & $+$0.51 & $+$0.44 & $-$0.05 & $-$0.07 \\
\bottomrule
\end{tabular}
}
\caption{Mean Likert score differences ($\Delta$) between human summaries and the five-model average for original references and contamination-free summaries. Positive values favor human summaries; negative values favor model summaries. \textit{Info.}\ = informativeness; \textit{Faith.}\ = faithfulness; \textit{Coher.}\ = coherence; \textit{Conci.}\ = conciseness.}
\label{tab:controlled_human}
\end{table*}

\autoref{tab:controlled_human} reports the score differences for the original human references and the contamination-free human summaries, broken down by dataset and dimension. For a fair comparison, the original reference summaries are recomputed on the same 24-sample subset used in the controlled condition. Under contamination-free conditions, the human advantage in informativeness and faithfulness widens on every dataset relative to the original references. Coherence and conciseness remain directionally consistent, with model summaries retaining a slight edge on coherence.

\section{Do LLM Judges Favor LLMs?}
\label{sec:judge_eval}

We employ all five evaluated models as LLM judges, each scoring summaries from all six sources across the same five datasets. Every model thus serves both as a candidate system and as a judge, enabling cross-model evaluation and self-preference analysis. Following the human-evaluation protocol, each judge assigns per-dimension Likert scores on informativeness, faithfulness, coherence, and conciseness, together with an overall ranking. To mitigate self-preference bias, each model's own judgment is excluded when scoring its summary, and pairwise win rates are derived from the averaged rankings of the remaining four judges. Before the evaluation, a pilot study (\autoref{appendix:judge_alignment}) provides a diagnostic check that judge scores are consistent with human evaluations, with no judge excluded or adjusted on the basis of its outcome, so the judge findings below constitute independent corroboration rather than calibration to the human signal.\footnote{Close judge--human alignment is not a prerequisite for the LLM-as-Judge track; the pilot only verifies that no judge exhibits categorically anomalous behavior before deployment.} Details on prompt templates, ranking protocol, and bias mitigation are in \autoref{appendix:judge_setup}.

\begin{figure*}[t]
  \centering
  \includegraphics[width=\textwidth]{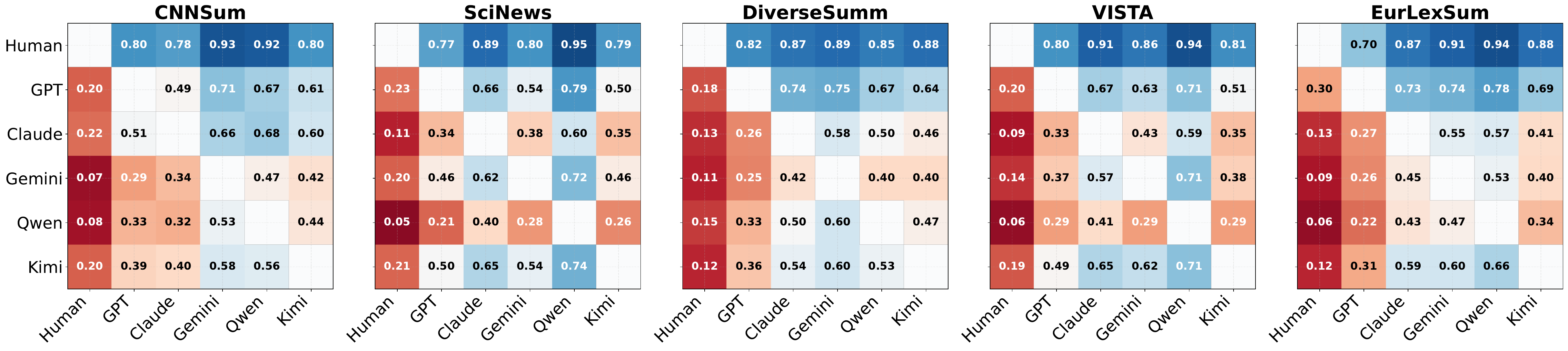}
  \caption{Pairwise win rates from LLM-as-Judge evaluation, shown separately for each dataset. Under the self-exclusion protocol, each cell reports the proportion of samples for which the row system is ranked above the column system, averaged across judges under the self-exclusion protocol within each sample.}
  \label{fig:judge_pairwise}
\end{figure*}

\begin{figure*}[t]
  \centering
  \includegraphics[width=\textwidth]{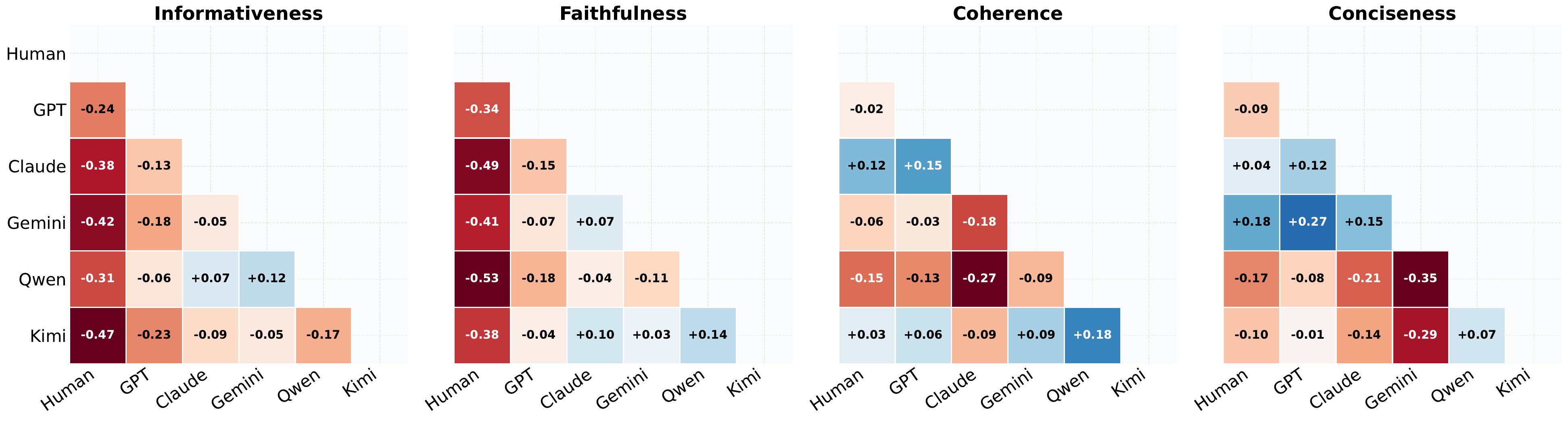}
  \caption{Likert score differences from LLM-as-Judge evaluation across four dimensions. Each lower-triangular cell reports the mean score of the row system minus that of the column system, averaged across datasets after first averaging judges under the self-exclusion protocol within each sample.}
  \label{fig:judge_agreement}
\end{figure*}

\autoref{fig:judge_pairwise} presents the pairwise win-rate matrices across datasets. Human summaries achieve win rates of 0.70 to 0.95 against all five models. These judge win rates are slightly more decisive than the crowd win rates of 0.66 to 0.90 in \S\ref{sec:human_eval}, consistent with bias-mitigated judges applying the per-dimension rubric more uniformly across samples than individual human raters. As in the human evaluation, model rankings vary across datasets, and no single model holds a stable position. \autoref{fig:judge_agreement} provides a dimension-level breakdown of score gaps. Human summaries again show positive gaps on informativeness and faithfulness, while the gaps on coherence and conciseness are reduced, with several models matching or slightly exceeding human summaries on these form-oriented dimensions.

\paragraph{Source-Length Stratification.} Longer documents place greater demands on information integration, which may affect human and LLM summarizers differently. Human writers can selectively draw on content from different sections of a long paper, whereas LLMs are known to lose coverage of information in the middle of long input sequences \citep{liu-etal-2024-lost}. To assess whether document length moderates the human--model gap, we partition the SciNews test samples into three equal tertiles by source document length and report the Human-minus-model-average difference ($\Delta$) on each of the four evaluation dimensions.

\begin{table*}[t]
\centering
\small
\setlength{\tabcolsep}{25pt}
\resizebox{\textwidth}{!}{%
\begin{tabular}{lcccc}
\toprule
Tertile & Info.\ $\Delta$ & Faith.\ $\Delta$ & Coher.\ $\Delta$ & Conci.\ $\Delta$ \\
\midrule
Short (0--33\%)  & $+$0.18 & $+$0.26 & $-$0.03 & $+$0.04 \\
Medium (34--66\%) & $+$0.29 & $+$0.35 & $+$0.05 & $+$0.06 \\
Long (67--100\%)   & $+$0.54 & $+$0.57 & $+$0.13 & $+$0.18 \\
\bottomrule
\end{tabular}
}
\caption{Mean Likert score differences ($\Delta$) between human summaries and the five-model average on SciNews, stratified by source-length tertile. Positive values favor human summaries; negative values favor model summaries.}
\label{tab:length_moderation}
\end{table*}

\autoref{tab:length_moderation} shows that the sign of the informativeness, faithfulness, and conciseness gaps is consistent with the previous SciNews findings, while coherence reverses from negative on the short tertile to positive on medium and long. The informativeness gap widens from $+$0.18 in the short tertile to $+$0.54 in the long tertile, and the faithfulness gap widens from $+$0.26 to $+$0.57. The coherence gap reverses from a slight model advantage on short documents ($-$0.03) to a clear human advantage on long ones ($+$0.13), indicating that the model's coherence edge does not survive the extra information-integration demand of longer inputs.

\section{Do Humans Hallucinate More Than LLMs?}
\label{sec:factuality}

We evaluate factual consistency with four complementary methods (FaStFact, SAFE, FActScore, and VeriScore), each of which decomposes summaries into (atomic) claims and verifies them against external knowledge sources (see \autoref{appendix:factuality_setup} for details). \autoref{fig:factuality_raincloud} presents the score distributions across the four metrics, five datasets, and six sources. Human summaries receive higher scores than model summaries, with average margins of 0.04 to 0.13. The ordering is consistent across all four methods despite their differences in claim granularity and evidence retrieval.

\begin{figure*}[t]
  \centering
  \includegraphics[width=\textwidth]{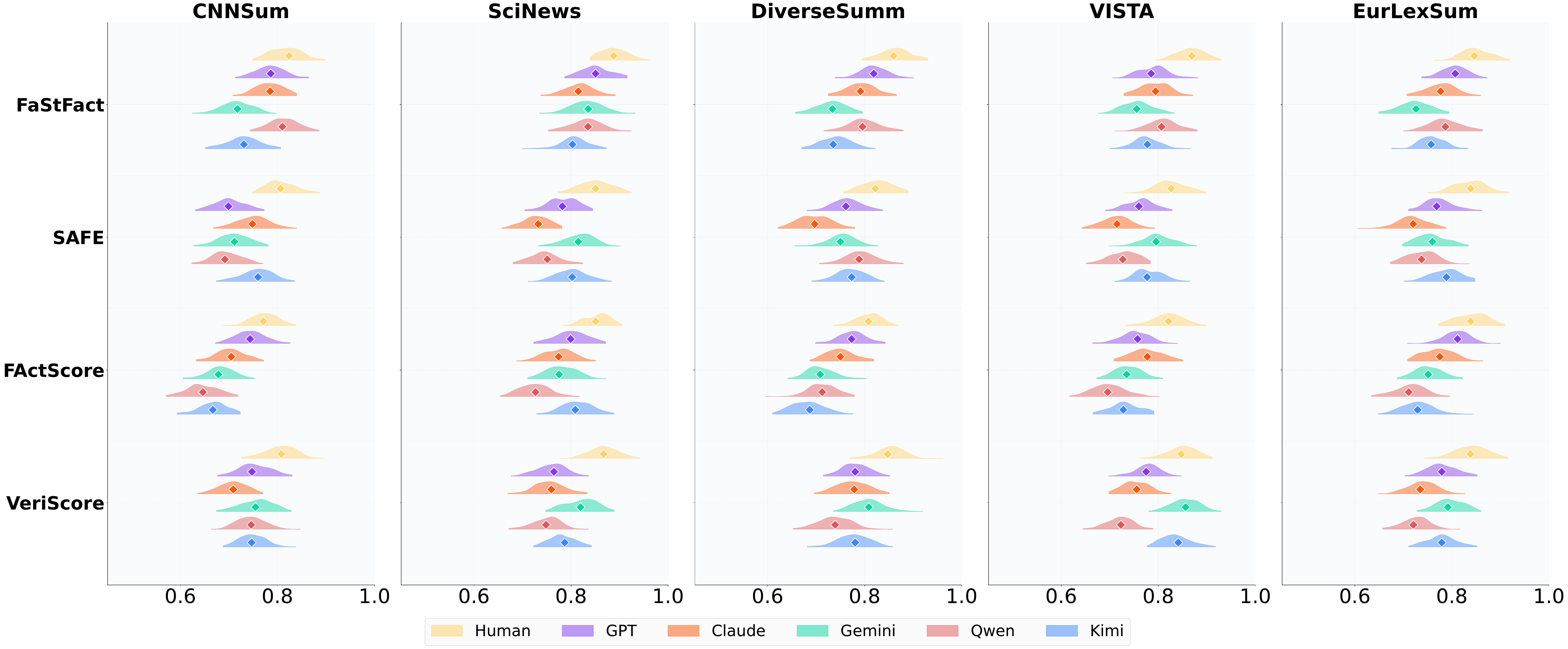}
  \caption{Factuality score distributions across four verification methods, five datasets, and six summary sources. Each half-violin shows one distribution, and diamonds mark the means.}
 \label{fig:factuality_raincloud}
\end{figure*}

No single model achieves the highest score across all four metrics simultaneously. These findings complement prior studies that reported LLM summaries as more factually consistent than human references \cite{tam-etal-2023-evaluating}. Such conclusions are typically reached under source-only verification, where any content that is not grounded in the source document is treated as hallucination. This operationalization is internally consistent, but it conflates two phenomena that differ in origin and evaluative consequence \cite{qi-etal-2025-evaluating}. 

A genuine/actual hallucination is a claim that cannot be verified against any credible knowledge source. What counts as one is itself prompt-dependent, and both human and LLM summaries can carry extrinsic content of either kind. In our setup (\autoref{appendix:prompts}), four prompts (CNNSum, DiverseSumm, VISTA, and EurLexSum) explicitly forbid introducing information absent from the source, while the SciNews prompt invites accessible explanation of domain-specific terms; the source-only ablation in \autoref{appendix:source_only_factuality} aligns with these instructions, with human references scoring below the five-model average only on SciNews under source-only verification and recovering once claims are evaluated against external knowledge \cite{tang-etal-2024-tofueval, dong-etal-2022-faithful, ramprasad-etal-2024-evaluating, rahman2026hallucination}. This asymmetry reflects a provenance-blind evaluation protocol that conflates purposeful contextual enrichment with unsupported generation. A case study illustration is provided in \autoref{appendix:case_study}, where the present paper serves as the source document and is free of data contamination.

\section{Do Human and LLM Summaries Differ Linguistically?}
\label{sec:linguistic_divergence}

\begin{figure*}[t]
  \centering
  \includegraphics[width=\linewidth]{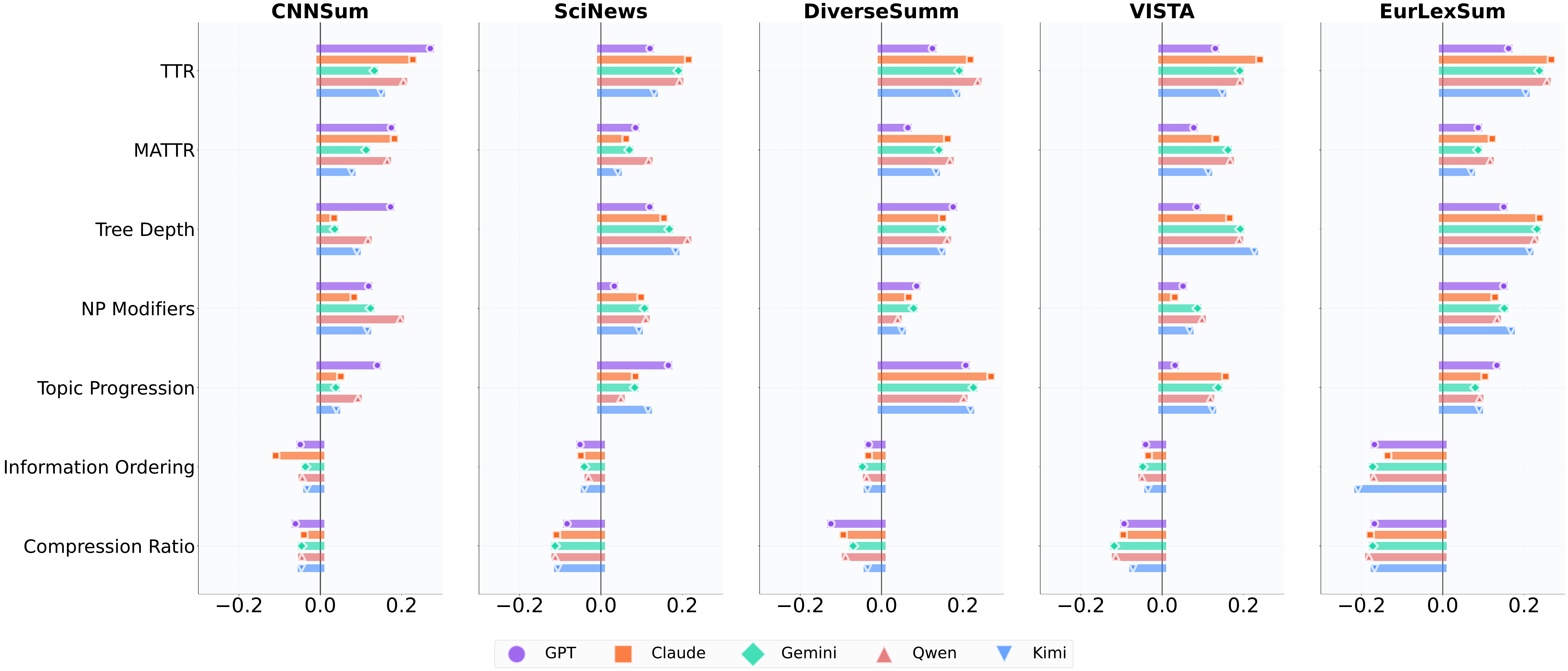}
  \caption{Linguistic differences between human and model summaries across seven metrics and five datasets. Values are reported as $\Delta = \text{Human} - \text{Model}$; positive values indicate higher human scores, and negative values indicate higher model scores.}
  \label{fig:dotplot_all_models}
\end{figure*}

We compare the linguistic properties of human and LLM summaries at three levels of analysis, namely lexical (word-level), syntactic (sentence-level), and discourse (document-level). Across these levels, we report seven metrics in three groups. TTR and MATTR \citep{Covington01052010} measure lexical diversity. Dependency tree depth (Tree Depth) and noun phrase modifier count (NP Modifiers) capture syntactic complexity. Topic progression, information ordering, and compression ratio characterize discourse-level properties \cite{nenkova-louis-2008-summarize, davoodijam2024evaluation}. All summaries are processed with \texttt{Stanza} \citep{qi2020stanza}, with \texttt{Qwen3-Embedding-8B} \citep{zhang2025qwen3} for sentence embeddings. Metric definitions, normalization procedures, and preprocessing details are provided in \autoref{appendix:linguistic_setup}.

\autoref{fig:dotplot_all_models} presents the divergence between human and model summaries across all metrics and datasets, reported as $\Delta = \text{Human} - \text{Model}$. Human summaries score higher on lexical and syntactic complexity, whereas model summaries adhere more closely to source ordering and apply less aggressive compression. We discuss each level in turn.

At the lexical level, human summaries exhibit higher TTR and MATTR values than model summaries across all datasets, with $\Delta$ values ranging from $+$0.08 to $+$0.28 for TTR and from $+$0.06 to $+$0.21 for MATTR. This phenomenon suggests that humans employ a more varied vocabulary, potentially reflecting a preference for paraphrasing and synonym substitution. Model summaries appear to favor the repeated use of domain-relevant terms, which may improve topical cohesion at the expense of surface-level lexical variety.

At the syntactic level, human summaries consistently produce deeper dependency trees ($\Delta$ of $+$0.05 to $+$0.24) and carry more modifiers per noun phrase ($\Delta$ of $+$0.04 to $+$0.19). These differences indicate that human writers construct more complex structures with richer noun phrase elaboration, whereas model summaries favor shallower trees and leaner noun phrases. This simpler syntactic profile may contribute to the perceived fluency reported in human evaluation studies, but at the cost of information density within individual sentences.

At the discourse level, models score higher on information ordering ($\Delta$ ranges from $-$0.03 to $-$0.19), indicating that models follow the linear order of the source document more closely than humans. Human writers appear more willing to reorganize content according to narrative or argumentative logic. Humans exhibit lower compression ratios ($\Delta$ ranges from $-$0.03 to $-$0.20), meaning that human summaries are shorter relative to the source and apply stronger compression. Topic progression shows positive $\Delta$ values ($+$0.03 to $+$0.26), suggesting that consecutive sentences in human summaries are more semantically similar.

These linguistic characteristics offer a structural account for the quality trade-offs observed in the preceding evaluation tracks. The shallower syntactic structures and lower lexical diversity of model summaries likely contribute to the surface fluency that annotators and LLM judges reward on the coherence dimension \cite{zhou2026fairnessfluencyinvestigationlanguage, ryu-etal-2024-multi}. Differences in information density and sentence-level packaging may also help explain why model summaries are perceived as concise despite not compressing the source more aggressively than human summaries. The stylistic uniformity across model families further suggests that these properties reflect shared tendencies in current LLM architectures or training regimes rather than model-specific idiosyncrasies. Statistical significance tests for the score differences across all four evaluation tracks are reported in \autoref{appendix:significance}.

\section{Why Do These Gaps Matter?}

Beyond serving as a research subject, summarization functions as an enabling capability woven into a variety of downstream NLP systems. When a summary omits key information, distorts facts, or applies rigid surface forms, those weaknesses propagate into the systems built on top of it. The gaps documented in the preceding evaluations have direct consequences in the pipelines surveyed below.

\paragraph{Retrieval and Knowledge Grounding.} In retrieval-augmented generation (RAG), summarization touches indexing, chunking, and re-ranking. Summary-based index entries can foreground salient content and reduce embedding noise; summary-guided segmentation produces more coherent retrieval units than fixed-length splitting; and query-focused summarization compresses candidate passages into more effective representations for relevance estimation \cite{wang-etal-2025-document, zhao-etal-2025-moc, edge2024local, wang2025archrag}. Across these stages, the summarization sets an implicit ceiling on retrieval effectiveness, and omissions or factual distortions propagate silently to the generation stage \cite{tamber-etal-2025-benchmarking, liu-etal-2026-disco}. As RAG sees broader adoption in knowledge-intensive domains, improving its summarization layer is likely to yield compounding returns.

\paragraph{AI-Powered Search and Question Answering.} Summarization is central to AI-powered search and question answering, where the demands on compression compound across use cases. Search snippet generation requires extracting the most informative portions of a web page under tight length constraints. Multi-document QA requires aggregating evidence across passages while resolving redundancy and reconciling potentially conflicting claims \cite{balepur-etal-2025-mods, zhang-etal-2025-belle}. Attributed generation, in which answers carry inline citations to supporting sources, adds another layer by demanding that provenance be preserved through compression \cite{wright-etal-2025-unstructured}. Shortcomings in the underlying summarization, whether through omission of key evidence, unfaithful compression, or difficulty in reconciling contradictions, translate directly into degraded answer quality.

\paragraph{Professional Workflows and Decision Support.} In professional settings, summarization mediates between large volumes of unstructured information and the structured outputs that inform decision-making. Video summarization goes beyond producing a shortened transcript; it involves extracting action items, logging decisions, and identifying follow-up tasks from fragmented exchanges \cite{10.1145/3711074}. In clinical environments, distilling patient encounters into electronic health records requires domain expertise to separate clinically relevant observations from incidental conversation \cite{nagar-etal-2025-umedsum}. Legal discovery condenses case documents into summaries that retain essential details \cite{malik-etal-2024-civilsum}, while financial analysis depends on summarizing earnings calls and regulatory filings, where omission of a single quantitative detail can shift interpretation \cite{li-etal-2025-legalagentbench}. In these domains, summarization shortcomings are amplified, and the gap between what current systems deliver and what such workflows demand underscores the continued need for domain-adapted, controllable, and faithful summarization \cite{pu-etal-2022-two}.

\section{Open Challenges}

The gaps documented across our evaluation tracks, and amplified through the downstream pipelines surveyed above, reflect methodological challenges that the field has not yet resolved, rather than artifacts of any single dataset, metric, or model choice. We highlight three open problems concerning factuality verification, summary evaluation, and benchmark credibility since these challenges are interconnected, as verification evidence determines which errors can be detected, evaluation protocols shape which qualities are rewarded, and benchmark design affects whether reported gains remain meaningful as models evolve.

\paragraph{Factuality Verification.} Ensuring that every claim in a generated summary is factually grounded remains a central concern \cite{maynez-etal-2020-faithfulness}. The community has made strides through NLI-based consistency metrics and model-based verification pipelines, which detect coarse-grained errors such as entity substitutions and negation flips \cite{zhao-etal-2020-reducing,utama-etal-2022-falsesum, scire-etal-2024-fenice}. However, LLM-generated summaries often contain subtle distortions, including quantity shifts, causal reversals, and temporal re-orderings. Such errors are common in claims that require reasoning or synthesis and remain difficult to detect reliably \cite{zha-etal-2023-alignscore}. The choice of verification protocol, whether restricted to the source document or augmented with external knowledge, can also influence conclusions about factual consistency \cite{cao-etal-2022-hallucinated}.

\paragraph{Evaluation Protocols.} The evaluation of summarization systems has become an active area of inquiry. Traditional reference-based metrics such as ROUGE and BERTScore align less well with human quality judgments for LLM-generated outputs, partly because they capture surface-level overlap and may fail to reflect broader differences in information selection and organization \cite{forde-etal-2024-evaluating}. Our findings imply that dimension-specific assessment can expose quality trade-offs that holistic judgments leave implicit \cite{song-etal-2024-finesure}. LLM-based evaluators in turn require careful bias mitigation, since design choices such as presentation order and the handling of self-preference can alter the resulting rankings \cite{koo-etal-2024-benchmarking}. Developing standardized, reproducible evaluation protocols remains a shared objective for the field.

\paragraph{Benchmark Design.} The quality of evaluation benchmarks has attracted growing scrutiny. Data contamination remains a significant concern, as widely used benchmarks such as \texttt{CNN/DailyMail} and \texttt{XSum} are likely included in the pretraining corpora of many LLMs \cite{golchin2024time}. Contamination is unavoidable and cannot be eliminated \cite{xu-etal-2025-dcr, sainz-etal-2023-nlp}. Existing benchmarks remain concentrated in English-language news and dialogue; findings derived from high-resource languages may not generalize to lower-resource settings \cite{urlana-etal-2023-pmindiasum, zhang-etal-2024-benchmarking, forde-etal-2024-evaluating}. A promising direction is the adoption of arena-style evaluation platforms with non-public and regularly refreshed test sets, which could mitigate contamination risks while enabling standardized comparison across systems.

\section{Conclusion}

We revisit the claim that LLMs have closed the gap with human references in summarization through a multi-track evaluation, combining dimension-specific human assessment, bias-mitigated LLM-as-Judge, external-knowledge factuality verification, and corpus-level linguistic analysis. Human summaries retain advantages in informativeness and faithfulness, while LLM outputs gain ground on the form-oriented dimensions of coherence and conciseness. Human summaries are factually more reliable, with their advantage becoming more pronounced once verification draws on external knowledge; LLM outputs remain systematically less diverse at the lexical, syntactic, and discourse levels across model families. Surface fluency does not entail information fidelity, and the dimensions on which human summaries excel are precisely those that single-track evaluations and surface-overlap metrics tend to miss. These gaps matter beyond benchmark rankings, since downstream systems inherit the omissions, distortions, and stylistic uniformity of the summaries on which they rely. Progress should be measured not only by whether outputs read fluently, but also by whether they preserve evidence, support reliable decisions, and adapt their compression to the demands of a task. We encourage the community to treat summarization not as a solved capability but as a continuing testbed for advances in faithful reasoning, information compression, and linguistically diverse generation.

\section*{Limitations}

\paragraph{Model Scope.} Our evaluation considers five general-purpose frontier LLMs accessed via API in a zero-shot setting. Given the pace of model updates and post-training refinements, the results are best interpreted as reflecting a snapshot of model behavior. Models fine-tuned for summarization or augmented with retrieval or iterative self-refinement may exhibit different quality profiles. To enhance coverage, we include models from multiple providers, although the extent to which the findings generalize to future releases remains a question for future work.

\paragraph{Domain Coverage.} Our datasets cover Chinese literary text (CNNSum), English popular science (SciNews), multi-document English news (DiverseSumm), academic talk videos (VISTA), and EU legal text in 24 official EU languages (EurLexSum), but do not include conversational summarization, code summarization, social-media text, clinical notes, or low-resource non-European languages such as Arabic, Hindi, or Swahili. The robustness of the directional patterns across these settings provides initial evidence that the conclusions are not idiosyncratic to a single domain or language family, although extending the evaluation to additional domains and language families remains a direction.

\paragraph{Data Contamination.} Data contamination is a widely discussed consideration in evaluations of LLMs. We select five datasets that span multiple domains and languages to reduce the likelihood of systematic overlap with pretraining corpora, although complete exclusion cannot be guaranteed. To probe the influence of contamination, we additionally construct newly written human summaries unlikely to appear in any model's training data and repeat the evaluation. Under this controlled condition, the advantage of human summaries becomes more pronounced, suggesting that any overlap in the original human references would attenuate rather than amplify the observed differences.

\paragraph{Human Evaluation Scale.} Scaling human evaluation remains an ongoing constraint. We employ larger annotator pools, stricter qualifications, and higher per-sample redundancy than most prior work. The per-dataset sample size still reflects a trade-off between coverage and annotation cost, which is a constraint common to studies that prioritize annotation quality. The consistency of findings across five datasets and their alignment with the LLM-as-Judge track support the directional conclusions reported in our study, while broader coverage could provide additional statistical support.

\paragraph{Annotator Heterogeneity.} Crowd annotators vary in education, domain expertise, and judgment style, and for the 24-language EurLexSum evaluation, each language's annotators bring their own legal and regulatory backgrounds that may shape interpretations of faithfulness and informativeness. We mitigate this through native-speaker recruitment for each language, calibration rounds prior to formal annotation, and continuous monitoring of Krippendorff's $\alpha$ within each track, although individual variation cannot be fully eliminated.

\paragraph{LLM-as-Judge Reliability.} The LLM judges we applied may share systematic biases inherited from common pretraining or preference-tuning regimes, and our self-exclusion protocol mitigates self-preference at the level of individual models but not such family-level effects. We probe this through a pilot study with human annotators (\autoref{appendix:judge_alignment}), and the convergence of the human-only and judge-only results on the same content-versus-form asymmetry provides the cross-check available within the scope of this study.

\paragraph{Verification Evidence Coverage.} Open-domain factuality verification relies on web search, and specialized content such as Chinese literary fiction or certain EU legal texts may be underrepresented in search indices, affecting factuality estimates for both human and model summaries. We mitigate this by employing four complementary verification methods with distinct reasoning strategies, so that the limitations of any single approach are partially balanced by the others. Incorporating domain-specific knowledge bases in future work may further improve coverage.

\paragraph{Linguistic Metric Coverage.} No finite metric set fully captures linguistic quality. Our seven metrics span three complementary levels (lexical, syntactic, and discourse), and the consistency of the findings across five datasets and 24 EurLexSum languages supports the conclusions. Properties such as pragmatic appropriateness, register consistency, and reader-perceived naturalness are not directly covered, and incorporating additional measures would offer a more comprehensive perspective.

\paragraph{Generation Configuration.} The main evaluation adopts greedy decoding, a fixed maximum token budget, and a per-dataset prompt template for controlled, reproducible comparison. Auxiliary experiments on prompt variation, decoding temperature, chain-of-thought prompting, and self-correction (\autoref{appendix:prompt_sensitivity}, \autoref{appendix:prompt_strategies}) show that the directional patterns remain stable. The configuration space is nonetheless large, and alternative strategies may yield different quality profiles. All models receive identical prompts per dataset, so any configuration-related effects apply uniformly rather than favoring a particular system.

\paragraph{Reference Quality.} The human reference summaries are taken from the original dataset releases and were produced under a range of conditions, including professional annotators, crowd workers, and domain experts. This variation is typical of summarization benchmarks and reflects the diversity of human summarization practices. While these references may not represent the upper bound of human performance, their consistent advantage over model outputs suggests that reference quality alone is unlikely to explain the observed differences.

\section*{Ethical Considerations}
This study draws on publicly released datasets, with all usage conforming to the respective licenses and distribution terms. Annotators participate voluntarily and receive appropriate compensation, with calibration rounds offered prior to formal annotation. Precautions are taken to reduce the likelihood that annotators encounter harmful content beyond what appears in the original corpora. The evaluation pipeline operates on de-identified data and does not collect or attempt to infer personally identifiable information. Large language models are used both as the subject of evaluation and as supporting components. The study is conducted in alignment with the \href{https://www.aclweb.org/adminwiki/index.php/ACL_Policy_on_Publication_Ethics}{ACL Policy on Publication Ethics}.

\bibliography{custom}

\appendix

\section{Model and Dataset Details}
\label{appendix:setup_details}

\autoref{tab:model_ids} lists the full model identifiers and API access dates for all five evaluated LLMs. Throughout the paper, we use abbreviated names for simplicity.

\begin{table}[h]
\centering
\small
\setlength{\tabcolsep}{15pt}
\resizebox{0.98\columnwidth}{!}{%
\begin{tabular}{ll}
\toprule
Abbreviation & Full Model Identifier \\
\midrule
GPT & \texttt{GPT-5.4-2026-03-05} \\
Claude & \texttt{Claude-opus-4-6} \\
Gemini & \texttt{Gemini-3.1-pro-preview} \\
Qwen & \texttt{Qwen3.5-397B-A17B} \\
Kimi & \texttt{Kimi-K2.5} \\
\bottomrule
\end{tabular}}
\caption{Abbreviations and full identifiers for the five evaluated LLMs. All models were accessed through their respective APIs between March 1 and May 15, 2026.}
\label{tab:model_ids}
\end{table}

Below, we describe each dataset used in our evaluation.

\begin{itemize}[leftmargin=10pt,itemsep=4pt,topsep=4pt]
\item CNNSum \citep{wei-etal-2025-cnnsum} is a Chinese novel summarization dataset of serialized web novels with long source documents; human-editor references condense entire story arcs into coherent synopses. We use the full training set (695 samples), as no official test/validation split is provided.

\item SciNews \citep{pu-etal-2024-scinews} is an English lay summarization dataset that pairs scientific research papers with news articles written for general audiences, requiring simplification of technical content while preserving accuracy. We use the official full test split (4,188 samples).

\item DiverseSumm \citep{huang-etal-2024-embrace} is an English multi-document news summarization dataset that aggregates multiple articles on the same event into a single coherent summary. We use all available samples from the training set (245), as no official test/validation split is provided.

\item VISTA \citep{liu-etal-2025-talk} is an English video-to-text summarization dataset pairing scientific paper abstracts with video presentations. We use the official full test split (1,859 samples).

\item EurLexSum \citep{aumiller-etal-2022-eur} is a multilingual legal summarization dataset of EU legislative documents and their official summaries in 24 EU languages, with parallel summaries enabling controlled cross-lingual comparison. We use the official full test split (188 samples per language; 4,512 in total).
\end{itemize}

\section{Human Evaluation}
\label{appendix:human_eval}

\subsection{Human Annotation Setup}
\label{appendix:human_eval_setup}

\paragraph{Expert Annotators.} We assemble five domain experts, all holding doctoral degrees in fields aligned with the datasets: Chinese literature (CNNSum), natural sciences (SciNews), journalism and media studies (DiverseSumm), computer science with a focus on artificial intelligence (VISTA), and EU law (EurLexSum).\footnote{Experts may use translation tools to assist comprehension of texts in the relevant language.} Each expert independently annotates a seed set of 20 samples from their area; for each sample, all candidate summaries are shown in blind, randomized order, and the expert assigns 1-to-5 Likert scores on the four dimensions (informativeness, faithfulness, coherence, conciseness) plus a listwise overall ranking (1 = worst to 6 = best). These annotations serve as (i) references for the crowdsourcing qualification test and (ii) calibration data for the annotation protocol before the main study.

\paragraph{Annotator Recruitment.} We recruit crowd annotators through Prolific (\url{https://www.prolific.com}). Eligibility requires (i) a prior-task approval rate of at least 85\%, (ii) at least an undergraduate-level education, and (iii) self-reported language proficiency: CEFR C1 or above in English for SciNews, DiverseSumm, and VISTA; native Chinese for CNNSum; and native-level proficiency in the respective EU language for EurLexSum.

\paragraph{Qualification Test.} Each candidate completes a qualification round on the same 20 expert-annotated seed samples, following the same protocol as the experts. We then compute Pearson $r$ between the candidate's and the experts' Likert scores per dimension across the 20 samples and require $r \geq 0.65$ on every dimension; candidates below the threshold are not admitted. Of 787 candidates who attempted the qualification, 167 passed and were retained.

\paragraph{Annotation Protocol.} The full guidelines are given in \autoref{fig:human_eval_guidelines}. Each sample follows a blind listwise protocol. The source document and all six candidate summaries appear simultaneously, with system identities hidden and presentation order randomized independently per annotator. For each summary, annotators assign 1-to-5 Likert scores on the four dimensions and produce a single overall listwise ranking from 1 (worst) to 6 (best). Each sample is independently evaluated by three crowd annotators.

\paragraph{Pairwise Win-Rate Derivation.} Pairwise win rates are derived from the overall listwise rankings. For each annotator and each pair (A, B), A wins if its rank is strictly better than B's, with ties split evenly. The win rate of A over B is the average across all annotators and samples, yielding a value in [0, 1] satisfying win\_rate(A, B) + win\_rate(B, A) = 1.00.

\paragraph{Quality Control.} 5\% of samples are duplicated throughout the annotation process to measure intra-annotator consistency. Inter-annotator agreement is monitored via Krippendorff's $\alpha$ every 20 completed samples, with recalibration reminders issued whenever $\alpha$ falls below 0.70. Annotators whose ongoing Pearson correlation with the remaining annotators drops below 0.60 are excluded from the final analysis.

\paragraph{Inter-Annotator Agreement.} \autoref{tab:iaa} reports Krippendorff's $\alpha$ (ordinal weighting) for each Likert dimension and dataset, computed on the final accepted annotations after quality-control exclusions. All values meet or exceed the 0.70 monitoring threshold.

\begin{table}[h]
\centering
\small
\setlength{\tabcolsep}{10pt}
\resizebox{0.9\linewidth}{!}{%
\begin{tabular}{lcccc}
\toprule
Dataset & Info. & Faith. & Coher. & Conci. \\
\midrule
CNNSum      & 0.84 & 0.74 & 0.76 & 0.73 \\
SciNews     & 0.76 & 0.74 & 0.72 & 0.76 \\
DiverseSumm & 0.75 & 0.78 & 0.74 & 0.70 \\
VISTA       & 0.78 & 0.72 & 0.73 & 0.71 \\
EurLexSum & 0.81 & 0.78 & 0.76 & 0.73 \\
\bottomrule
\end{tabular}
}
\caption{Inter-annotator agreement by dataset and evaluation dimension, measured using ordinal Krippendorff's $\alpha$. \textit{Info.}\ = informativeness; \textit{Faith.}\ = faithfulness; \textit{Coher.}\ = coherence; \textit{Conci.}\ = conciseness.}
\label{tab:iaa}
\end{table}

\subsection{Prompt and Temperature Sensitivity}
\label{appendix:prompt_sensitivity}

We generate all summaries using a domain-specific instruction template and greedy decoding ($T = 0$). To verify that the main findings are robust to these design choices, we examine sensitivity along two consequential generation factors, namely the level of prompt specificity and the decoding temperature.

\paragraph{Prompt Conditions.} We compare two prompt variants applied identically to all five models on all five datasets. The first is the prompt from \autoref{appendix:prompts}, which provides domain-specific guidance, explicit quality criteria, and an output-format instruction (hereafter \textit{Default}). The second is a \textit{Minimal} prompt consisting of a single-sentence instruction to summarize the source document, with no domain guidance or format specification.

\paragraph{Temperature Conditions.} Using the default prompt, we additionally generate summaries at $T = 0.3$ and $T = 0.7$ under nucleus sampling (top-$p = 0.95$) and compare them with the greedy baseline ($T = 0$).

\paragraph{Evaluation Protocol.} For each condition, we run the same human evaluation on a held-out subset of 24 samples per dataset (120 samples total), using the same annotation protocol and annotator pool as the main evaluation.\footnote{One sample is drawn from each language in EurLexSum.} We report the average pairwise win rate of human summaries against the five-model average on each of the four evaluation dimensions.

\begin{table*}[t]
\centering
\small
\setlength{\tabcolsep}{25pt}
\resizebox{0.95\textwidth}{!}{%
\begin{tabular}{lcccc}
\toprule
Condition & Info. & Faith. & Coher. & Conci. \\
\midrule
\multicolumn{5}{l}{\textit{Prompt condition (decoding: greedy $T = 0$)}} \\
\midrule
Minimal instruction  & 0.75 & 0.74 & 0.50 & 0.53 \\
\textbf{Default (ours)} & \textbf{0.71} & \textbf{0.68} & \textbf{0.42} & \textbf{0.49}\\
\midrule
\multicolumn{5}{l}{\textit{Decoding temperature (prompt: Default)}} \\
\midrule
\textbf{Greedy $T = 0$ (ours)} & \textbf{0.71} & \textbf{0.68} & \textbf{0.42} & \textbf{0.49} \\
Sampling $T = 0.3$ & 0.70 & 0.69 & 0.45 & 0.51 \\
Sampling $T = 0.7$ & 0.70 & 0.69 & 0.42 & 0.48 \\
\bottomrule
\end{tabular}
}
\caption{Mean pairwise win rates of human summaries against the five models under variations in prompt specificity and decoding temperature. Results are averaged across five datasets on a held-out set of 120 samples. \textbf{Bold} marks the settings used in the main evaluation; values above 0.50 favor human summaries, and values below 0.50 favor model summaries.}
\label{tab:prompt_sensitivity}
\end{table*}

The results are shown in \autoref{tab:prompt_sensitivity}. The human advantage on informativeness and faithfulness persists across every condition, with win rates spanning 0.70--0.75 and 0.68--0.74, respectively. Coherence stays at or below 0.50 throughout, confirming that models are preferred on this dimension regardless of prompt or temperature choice. The numerical differences across conditions are small, and no condition reverses any finding reported in the main text.

Within the prompt conditions, the minimal instruction increases human win rates on informativeness (0.71 to 0.75) and faithfulness (0.68 to 0.74). Within the temperature conditions, $T = 0.3$ has a slightly larger effect, raising coherence to 0.45 and conciseness to 0.51.

\subsection{Prompting Strategy Sensitivity}
\label{appendix:prompt_strategies}

We additionally examine two prompting strategies that alter the reasoning process underlying summary generation, namely chain-of-thought prompting and a two-stage self-correction procedure.

\paragraph{Chain-of-Thought Conditions.} We augment the prompt from \autoref{appendix:prompts} with a structured reasoning preamble that instructs each model to enumerate the three to five most important claims in the source document, identify supporting evidence for each, and then compose the summary from that claim list. The reasoning trace is discarded from the final output; only the composed summary is submitted for evaluation. This condition tests whether explicit intermediate reasoning yields more thorough factual grounding than direct generation.

\paragraph{Self-Correction Conditions.} We employ a two-stage procedure. In the first stage, each model generates a draft summary. In the second stage, the same model receives the source document together with its draft and is instructed to identify factual inaccuracies, omitted key claims, and unjustified statements, then produce a revised summary that addresses these issues. The second-stage prompt is applied without modification across all five models, and only the revised summary is evaluated.

\paragraph{Evaluation Protocol.} Both conditions are evaluated on the same 120-sample held-out set as \autoref{appendix:prompt_sensitivity}, using the identical annotation protocol and annotator pool. We report the average pairwise win rate of human summaries against the five-model average on each of the four evaluation dimensions.

\begin{table*}[t]
\centering
\small
\setlength{\tabcolsep}{25pt}
\resizebox{\textwidth}{!}{%
\begin{tabular}{lcccc}
\toprule
Strategy & Info. & Faith. & Coher. & Conci. \\
\midrule
\textbf{Default (ours)} & \textbf{0.71} & \textbf{0.68} & \textbf{0.42} & \textbf{0.49} \\
Chain-of-thought                & 0.66 & 0.65 & 0.42 & 0.54 \\
Self-correction                 & 0.69 & 0.64 & 0.44 & 0.56 \\
\bottomrule
\end{tabular}
}
\caption{Mean pairwise win rates of human summaries against the five models under three prompting strategies. Results are averaged across five datasets on a held-out set of 120 samples. \textbf{Bold} marks the setting used in the main evaluation; values above 0.50 favor human summaries, and values below 0.50 favor model summaries.}
\label{tab:prompt_strategies}
\end{table*}

The results are shown in \autoref{tab:prompt_strategies}. Chain-of-thought prompting narrows the human faithfulness advantage relative to the baseline (0.68 to 0.65), consistent with claim enumeration pushing models to ground their outputs more explicitly in the source before composition. The informativeness win rate decreases from 0.71 to 0.66. Self-correction reduces both faithfulness (0.64) and informativeness (0.69), plausibly because the revision stage tends to drop content the model judges as uncertain, occasionally at the cost of coverage. Coherence (0.44) and conciseness (0.56) win rates rise under self-correction. Across all three strategies, the human advantage on informativeness and faithfulness is preserved, with win rates above 0.50 throughout.

\section{LLM-as-Judge}
\subsection{LLM-as-Judge Setup}
\label{appendix:judge_setup}

The full prompt template for LLM-as-Judge is shown in \autoref{fig:judge_prompt}. For each sample, the source document and all six candidate summaries are presented to the LLM judge in a single prompt. The judge assigns a 1-to-5 Likert score to each summary on each of four dimensions and produces a listwise ranking of the six summaries from worst (rank 1) to best (rank 6).

To mitigate position bias, the presentation order of the six summaries is randomized independently for each sample and each judge, and system identities are hidden (summaries are labeled Summary A through Summary F). Each sample is evaluated by all five judges (GPT, Claude, Gemini, Qwen, Kimi). The final per-dimension Likert score and overall rank for each system on each sample are the arithmetic mean of the judges' assignments (four judges for self-exclusion). Self-preference bias, where a judge systematically assigns higher scores and ranks to its own outputs than the remaining judges do, is a known concern in LLM-based evaluation \citep{panickssery2024llm}. To address this, our primary results exclude self-judgments. When computing scores and rankings for a summary generated by model $M$, both the Likert scores and the overall ranking provided by $M$ are removed, and the remaining four judges' outputs are averaged.

We derive pairwise win rates from the listwise rankings. To be specific, for each pair (A, B), the win rate of A over B is the proportion of samples on which A's average rank is strictly better than B's, with ties (identical average ranks) split evenly. All judge calls use greedy decoding (Temperature $T = 0$) with a maximum output length of 1{,}024 tokens for reproducibility.

\subsection{Judge--Human Alignment Pilot}
\label{appendix:judge_alignment}

Before the full evaluation, we verify that each candidate judge produces per-dimension Likert scores and overall rankings consistent with human judgments. The verification draws on the same samples assessed by crowd annotators in the human evaluation track (\S\ref{sec:human_eval}); judges receive no feedback from this check, and no judge-specific adjustments are made.

\paragraph{Data.} We use all samples per dataset from the human-evaluated pool, each accompanied by per-dimension Likert scores and an overall listwise ranking from the crowd annotators.

\paragraph{Procedure.} Each of the five candidate judges (GPT, Claude, Gemini, Qwen, Kimi) receives the same anonymized prompt, randomization scheme, and greedy decoding settings as in the full evaluation (\autoref{appendix:judge_setup}). For each sample, each judge assigns 1-to-5 Likert scores on the four dimensions and produces a listwise ranking. We measure alignment to human judgments with two statistics. Kendall's $\tau$ is the rank correlation between the judge's listwise ordering and the human consensus ordering (averaged across annotators), computed per sample and then averaged across samples. Pairwise agreement is the proportion of the $\binom{6}{2} = 15$ system pairs on which the judge's pairwise preference (derived from its listwise ranking) matches the human majority preference, averaged across samples. Both statistics are reported per dimension in \autoref{tab:judge_human_alignment}. The judge's per-dimension ranking is obtained by sorting the six summaries by their Likert scores on that dimension (ties broken by average rank), and the human consensus ranking is obtained analogously from the averaged annotator scores.

\paragraph{Results.} \autoref{tab:judge_human_alignment} reports the alignment statistics. Kendall's $\tau$ ranges from 0.61 to 0.78 across judges and dimensions, and pairwise agreement ranges from 71.5\% to 84.2\%. We find no systematic ordering errors or dimension conflation in any judge and proceed with the full evaluation using the prompt as designed, with no modifications based on the alignment check.

\begin{table}[h]
\centering
\small
\setlength{\tabcolsep}{15pt}
\resizebox{0.96\linewidth}{!}{%
\begin{tabular}{lcccc}
\toprule
Judge & Info. & Faith. & Coher. & Conci. \\
\midrule
\multicolumn{5}{c}{\textit{Kendall's $\tau$}} \\
\midrule
GPT       & 0.78 & 0.70 & 0.72 & 0.68 \\
Claude    & 0.76 & 0.74 & 0.70 & 0.76 \\
Gemini    & 0.74 & 0.72 & 0.78 & 0.65 \\
Qwen      & 0.77 & 0.73 & 0.74 & 0.62 \\
Kimi      & 0.72 & 0.66 & 0.73 & 0.61 \\
\midrule
\multicolumn{5}{c}{\textit{Pairwise agreement (\%)}} \\
\midrule
GPT       & 84.2 & 82.5 & 78.3 & 75.1 \\
Claude    & 83.0 & 81.7 & 76.8 & 77.6 \\
Gemini    & 81.5 & 73.4 & 75.2 & 73.0 \\
Qwen      & 78.8 & 77.5 & 72.1 & 74.5 \\
Kimi      & 79.2 & 76.8 & 71.5 & 72.2 \\
\bottomrule
\end{tabular}
}
\caption{Judge--human alignment on the human-evaluated sample pool, measured using Kendall's $\tau$ and pairwise agreement (\%). \textit{Info.}\ = informativeness; \textit{Faith.}\ = faithfulness; \textit{Coher.}\ = coherence; \textit{Conci.}\ = conciseness.}
\label{tab:judge_human_alignment}
\end{table}

\section{Factuality Verification}
\subsection{Factuality Verification Setup}
\label{appendix:factuality_setup}

We employ four factuality evaluation methods. They share a common paradigm of decomposing summaries into atomic claims and verifying each claim against evidence, but they differ in decomposition granularity, evidence sources, and verification mechanisms. We describe each method and its key hyperparameters below.

\begin{itemize}[leftmargin=10pt,itemsep=4pt,topsep=4pt]
\item FaStFact \citep{wan-etal-2025-fastfact} adopts a two-stage strategy. In the first stage, the summary is passed to the claim extractor using the official \texttt{stride=0} setting, and atomic claims are extracted from this input. A confidence-based pre-verification step (threshold 0.995) labels high-confidence claims directly to reduce unnecessary evidence retrieval, while low-confidence or uncertain claims are sent to the second stage for retrieval and verification. In the second stage, the remaining claims are verified against document-level evidence collected by crawling full web pages retrieved via the Jina search/reader pipeline. The final score is the proportion of supported claims among all extracted claims.

\item SAFE \citep{wei2024longform} follows a four-step process. The summary is first decomposed into individual atomic facts by an LLM. Each fact is then made self-contained through decontextualization, which resolves pronouns and context-dependent references. A relevance filter removes facts unrelated to the original query. Finally, each remaining relevant fact is evaluated through up to five Google Search steps and labeled as supported or not supported. The score is the proportion of supported facts among those judged either supported or unsupported, with irrelevant facts excluded.

\item FActScore \citep{min-etal-2023-factscore} decomposes a generated text into atomic facts, each representing a single piece of information, and assigns a binary label to each fact according to whether it is supported by a reliable knowledge source. The labeling is performed by an automated retrieval-based estimator. The final score is the proportion of supported facts among all extracted facts.

\item VeriScore \citep{song-etal-2024-veriscore} filters out unverifiable content (opinions, metaphors, hypothetical statements) before scoring. Its pipeline has three steps. In claim extraction, the LLM identifies only verifiable factual claims from the summary. Evidence retrieval is performed via Google Search (Serper API, top-5 results per claim). Claim verification distinguishes supported, contradicted, and inconclusive claims, while the automatic scorer uses binary supported-versus-unsupported classification.
\end{itemize}

All four methods are run using their original, publicly released implementations and default prompt templates. We make two modifications to the default configuration. First, we unify the LLM backend across all methods to GPT-5.4 (\texttt{GPT-5.4-2026-03-05}) for claim decomposition, decontextualization, relevance checking, and verification. Using a single backend eliminates confounds from differences in decomposition or verification quality across LLMs, and GPT-5.4 provides the multilingual capability needed because EurLexSum spans 24 languages and CNNSum is in Chinese. Second, we expand the evidence pool to include the original source document alongside the web-retrieved external pages. Under our default protocol, every claim is therefore verified against the union of two evidence streams, namely the original source document and the external web evidence retrieved by the method. For the VISTA dataset, where the source material consists of video presentations, we use the converted text transcript as the source input.

\subsection{Source-Only Verification Ablation}
\label{appendix:source_only_factuality}

Our default protocol (\autoref{appendix:factuality_setup}) verifies each claim against both the original source document and the web-retrieved external evidence. Prior work on factuality in summarization has commonly relied on a source-only protocol, where each claim is verified against the original document rather than external knowledge. To situate our main findings within this methodological context, we re-run all four methods under a source-only configuration that disables web retrieval and restricts evidence to the source document, while all other steps (claim decomposition, decontextualization, and verification) remain identical to the default protocol.

\autoref{tab:source_only_factuality} reports factuality scores under source-only verification, averaged across FaStFact, SAFE, FActScore, and VeriScore, for each source and dataset. $\Delta$ is the Human score minus the average of the five model scores; positive values indicate that human summaries score higher.

\begin{table*}[t]
\centering
\small
\setlength{\tabcolsep}{20pt}
\resizebox{\textwidth}{!}{%
\begin{tabular}{lccccccr}
\toprule
Dataset & Human & GPT & Claude & Gemini & Qwen & Kimi & $\Delta$ \\
\midrule
CNNSum      & \textbf{0.843} & 0.789 & 0.782 & 0.771 & 0.776 & 0.781 & $+$0.063 \\
SciNews     & 0.689          & \textbf{0.752} & 0.748 & 0.751 & 0.729 & 0.724 & $-$0.052 \\
DiverseSumm & \textbf{0.816} & 0.763 & 0.758 & 0.745 & 0.753 & 0.737 & $+$0.065 \\
VISTA       & \textbf{0.798} & 0.748 & 0.743 & 0.731 & 0.726 & 0.721 & $+$0.064 \\
EurLexSum   & \textbf{0.850} & 0.783 & 0.803 & 0.787 & 0.784 & 0.779 & $+$0.063 \\
\bottomrule
\end{tabular}
}
\caption{Factuality scores under source-only verification, averaged across FaStFact, SAFE, FActScore, and VeriScore. Web retrieval is disabled, and evidence is restricted to the source document. \textbf{Bold} marks the highest score in each row; $\Delta$ is the human score minus the five-model average, so positive values favor human summaries.}
\label{tab:source_only_factuality}
\end{table*}

Human summaries retain higher scores on four of the five datasets (CNNSum, DiverseSumm, VISTA, and EurLexSum), with $\Delta$ ranging from $+$0.063 to $+$0.065. SciNews is the sole exception. Under source-only verification, all five models score above the human references, with the human score falling below the five-model average by 0.052. This reversal is consistent with the nature of the task. SciNews pairs research papers with lay summaries written for a general audience, and human writers routinely contextualize scientific findings by drawing on domain knowledge that, while factually accurate, is absent from the source paper. Under source-only verification, such content is treated as unsupported regardless of its accuracy, which reduces the measured factuality of human references relative to model outputs. LLM summaries tend to follow the source document more closely, as independently reflected in the higher information ordering scores reported in \S\ref{sec:linguistic_divergence}, and are therefore less affected by the source-only constraint. Once external web evidence is reintroduced under the default protocol, the human factuality score on SciNews recovers to a level comparable to that observed on the other four datasets, confirming that the source-only result primarily reflects the penalization of legitimate background enrichment rather than genuine inaccuracy. On CNNSum, DiverseSumm, VISTA, and EurLexSum, where summary content that departs from the source is less consistently attributable to background enrichment, the source-only results remain consistent with the main findings.

\section{Linguistic Analysis}
\label{appendix:linguistic_setup}

All linguistic analyses use the \texttt{Stanza} NLP toolkit \citep{qi2020stanza}, which provides trained pipelines for tokenization, part-of-speech tagging, lemmatization, and dependency parsing. From the resulting Universal Dependencies annotations, we derive four token- and syntax-level metrics. TTR is the ratio of unique lowercased tokens to total tokens, excluding tokens tagged PUNCT, SYM, or X. MATTR \citep{Covington01052010} averages TTR over a sliding window of 50 tokens; texts shorter than 50 tokens default to standard TTR. Tree depth is the per-sentence maximum depth of the head-derived dependency tree (computed via depth-first search), averaged across sentences. NP modifiers, computed for each NOUN or PROPN token, count dependents whose relation type belongs to \{amod, nmod, nummod, det, compound, flat, appos\}, averaged across all nouns in the text.

For the discourse-level metrics, topic progression splits the summary into sentences using regex-based rules. Chinese summaries are segmented at periods, exclamation marks, question marks, and semicolons, while summaries in other languages are segmented at sentence-final punctuation followed by whitespace. Each sentence is then encoded with \texttt{Qwen3-Embedding-8B} \citep{zhang2025qwen3}, and topic progression is reported as the cosine similarity between consecutive sentence embeddings. Information ordering splits the source document into overlapping chunks (56 characters for Chinese, 128 for other languages, 50\% overlap). Each summary sentence is aligned to the best-matching chunk by unigram overlap, and Kendall's $\tau$ is computed between the resulting position sequence and the ideal monotonic sequence. The compression ratio is the character count of the summary divided by that of the source.

\section{Statistical Significance Tests}
\label{appendix:significance}

For each of the four evaluation tracks, we assess whether the observed differences between human and model summaries are statistically significant at the sample level. We apply the Wilcoxon signed-rank test \citep{wilcoxon1945individual} to paired observations, where each pair consists of the human and model scores for the same source document. The test is non-parametric and does not assume normally distributed paired differences, which makes it suitable for the sample-level comparisons used throughout the paper. To control the false discovery rate across the multiple human-versus-model comparisons within each track, we apply the Benjamini--Hochberg procedure \citep{benjamini1995controlling} at a nominal level of $\alpha = 0.05$. Scores within each track are pooled across the five datasets before testing. \autoref{tab:significance} summarizes the outcomes, with $**$ marking $p < 0.01$ after correction, $*$ marking $0.01 \leq p < 0.05$ after correction, and n.s.\ marking $p \geq 0.05$.

\begin{table*}[t]
\centering
\small
\setlength{\tabcolsep}{20pt}
\resizebox{\textwidth}{!}{%
\begin{tabular}{llccccc}
\toprule
Track & Dimension / Metric & GPT & Claude & Gemini & Qwen & Kimi \\
\midrule
\multirow{4}{*}{Human Evaluation}
  & Informativeness    & $**$ & $**$ & $**$ & $**$ & $**$ \\
  & Faithfulness       & $**$ & $**$ & $**$ & $**$ & $**$ \\
  & Coherence          & $*$ & $**$ & $*$ & n.s. & $*$ \\
  & Conciseness        & n.s. & n.s. & $**$  & n.s. & n.s. \\
\midrule
\multirow{4}{*}{LLM-as-Judge}
  & Informativeness    & $**$ & $**$ & $**$ & $**$ & $**$ \\
  & Faithfulness       & $**$ & $**$ & $**$ & $**$ & $**$ \\
  & Coherence          & n.s. & $*$ & n.s. & $**$ & n.s. \\
  & Conciseness        & n.s. & n.s. & $**$ & $**$ & n.s. \\
\midrule
\multirow{4}{*}{Factuality}
  & FaStFact           & $**$ & $**$ & $**$ & $**$ & $**$ \\
  & SAFE               & $**$ & $**$ & $**$ & $**$ & $**$ \\
  & FActScore          & $**$ & $**$ & $**$ & $**$ & $**$ \\
  & VeriScore          & $**$ & $**$ & $**$ & $**$ & $**$ \\
\midrule
\multirow{7}{*}{Linguistic Analysis}
  & TTR                & $**$ & $**$ & $**$ & $**$ & $**$ \\
  & MATTR              & $**$ & $**$ & $**$ & $**$ & $**$ \\
  & Tree Depth         & $**$ & $**$ & $**$ & $**$ & $**$ \\
  & NP Modifiers       & $**$ & $**$ & $**$ & $**$ & $**$ \\
  & Topic Progression  & $**$  & $**$ & $**$ & $**$ & $**$ \\
  & Information Ordering & $**$ & $**$ & $**$ & $**$ & $**$ \\
  & Compression Ratio  & $**$ & $**$ & $**$ & $**$ & $**$ \\
\bottomrule
\end{tabular}
}
\caption{Statistical significance of human--model score differences across the four evaluation tracks. Wilcoxon signed-rank tests use Benjamini--Hochberg FDR correction ($\alpha = 0.05$) after pooling scores across the five datasets within each track; LLM-as-Judge tests follow the self-exclusion protocol. After correction, $**$ denotes $p < 0.01$, $*$ denotes $0.01 \leq p < 0.05$, and n.s.\ denotes $p \geq 0.05$.}
\label{tab:significance}
\end{table*}

\section{Case Study: Summarizing This Paper}
\label{appendix:case_study}

We submit the full text of the present paper to the five evaluated models and ask each to generate an abstract using the prompt shown in \autoref{fig:prompt_case_study}. Because the paper had not been publicly released at the time of model training, this example is free of data contamination. The human reference (\autoref{fig:case_human}) is the abstract written by the paper's authors, and the model-generated summaries from GPT, Claude, Gemini, Qwen, and Kimi are shown in \autoref{fig:case_gpt}, \autoref{fig:case_claude}, \autoref{fig:case_gemini}, \autoref{fig:case_qwen}, and \autoref{fig:case_kimi}, respectively. Errors in the model-generated summaries are identified by cross-checking each claim against the paper text and are typeset in \textcolor{red}{\uline{red with underline}} with letter labels; \autoref{tab:case_study_errors} catalogs each error together with the correct information.

\begin{figure*}[t]
\centering
\begin{tcolorbox}[width=\linewidth,
  colback=gray!5!white,
  colframe=gray!15!black,
  coltitle=black,
  colbacktitle=gray!25!white,
  fonttitle=\bfseries,
  title=Case Study Prompt,
]
You are an expert academic researcher. Your task is to write an abstract for the following research paper. \\
\textbf{Instructions:} \\
1. Summarize the paper's motivation, methodology, key findings, and conclusions. \\
2. Preserve the technical terminology and quantitative details reported in the paper. \\
3. Do not paraphrase quantities in a way that changes their precision or meaning. \\
4. Do not introduce interpretations or claims that go beyond what the paper explicitly states. \\
5. Write 100 to 200 words in clear, grammatically correct academic English. \\
\textbf{Paper:} \texttt{\{document\}}
\end{tcolorbox}
\caption{Abstract-generation prompt used in the contamination-free case study.}
\label{fig:prompt_case_study}
\end{figure*}

\begin{figure*}[t]
\centering
\begin{tcolorbox}[width=\linewidth,
  colback=gray!8!white,
  colframe=gray!45!black,
  coltitle=black,
  colbacktitle=gray!25!white,
  fonttitle=\bfseries,
  title=Human-Written Summary,
]
The progress of large language models (LLMs) has fueled claims that model-generated summaries rival or even surpass human-written references, raising questions about whether summarization remains an open research problem. We re-examine this narrative through a multi-track evaluation covering diverse datasets and state-of-the-art LLMs, combining controlled human assessment, bias-mitigated LLM-as-Judge protocols, factuality verification against external knowledge, and corpus-level linguistic analysis. Our findings reveal a more nuanced landscape in which human references continue to demonstrate advantages in informativeness and faithfulness, whereas LLM outputs are preferred mainly for surface-level coherence and fluency. Factuality verification indicates that human references remain more reliable, particularly for claims involving reasoning or synthesis, and linguistic analysis uncovers a pattern of stylistic homogeneity across different models. These observations suggest that current LLMs have raised the floor of summarization quality, but the ceiling of their performance remains below human capabilities.
\end{tcolorbox}
\caption{Human-written reference summary: the original abstract of this paper.}
\label{fig:case_human}
\end{figure*}

\begin{figure*}[t]
\centering
\begin{tcolorbox}[width=\linewidth,
  colback=blue!5!white,
  colframe=blue!35!black,
  coltitle=black,
  colbacktitle=blue!15!white,
  fonttitle=\bfseries,
  title=GPT-Generated Summary,
]
This paper challenges the prevailing narrative that large language models have largely addressed summarization, presenting a multi-track evaluation across five datasets and five frontier LLMs. The study employs dimension-specific human annotation, bias-mitigated LLM-as-Judge scoring, factuality verification against external knowledge sources, and systematic linguistic analysis. Results show that human summaries retain measurable advantages on informativeness and faithfulness. \textcolor{red}{\uline{Among the five evaluated models, Claude shows the smallest gap relative to human references in the human evaluation track, followed by GPT.}}\textsuperscript{(\textit{a})} The evaluation \textcolor{red}{\uline{controls for verbosity bias and self-preference bias through output-length normalization and self-judgment exclusion,}}\textsuperscript{(\textit{b})} and the resulting preference patterns differ from those obtained without such controls. Factuality analysis reveals that errors are concentrated on claims requiring reasoning or cross-sentence synthesis. Linguistic profiling confirms lower lexical diversity and shallower dependency trees in model outputs, suggesting a persistent gap in information density and linguistic richness.
\end{tcolorbox}
\caption{GPT-generated summary. Errors (\textit{a}) and (\textit{b}) are marked in \textcolor{red}{\uline{red with underline}}.}
\label{fig:case_gpt}
\end{figure*}

\begin{figure*}[t]
\centering
\begin{tcolorbox}[width=\linewidth,
  colback=green!4!white,
  colframe=green!40!black,
  coltitle=black,
  colbacktitle=green!15!white,
  fonttitle=\bfseries,
  title=Claude-Generated Summary,
]
This position paper argues that summarization remains an open research problem despite the strong performance of recent LLMs. The paper designs a multi-track evaluation framework covering five datasets and five LLMs, applying human annotation, LLM-as-Judge scoring, factuality verification, and linguistic analysis. Human annotators assess summaries on four dimensions, and \textcolor{red}{\uline{inter-annotator agreement is monitored via Cohen's kappa, with a threshold of 0.7 required for inclusion.}}\textsuperscript{(\textit{c})} Human summaries retain advantages in informativeness and faithfulness, while LLM outputs are preferred in coherence and conciseness. Linguistic analysis employs \textcolor{red}{\uline{eight metrics organized at three levels of analysis (lexical, syntactic, and discourse),}}\textsuperscript{(\textit{d})} revealing lower lexical diversity and shallower dependency structures in model outputs. The authors contend that while LLMs have narrowed the performance gap on surface form dimensions, persistent deficits in information density and factual reliability indicate that summarization is not yet solved.
\end{tcolorbox}
\caption{Claude-generated summary. Errors (\textit{c}) and (\textit{d}) are marked in \textcolor{red}{\uline{red with underline}}.}
\label{fig:case_claude}
\end{figure*}

\begin{figure*}[t]
\centering
\begin{tcolorbox}[width=\linewidth,
  colback=orange!5!white,
  colframe=orange!45!black,
  coltitle=black,
  colbacktitle=orange!18!white,
  fonttitle=\bfseries,
  title=Gemini-Generated Summary,
]
The paper presents a multi-dimensional evaluation of LLM-generated summaries against human references across five datasets covering Chinese fiction, lay science communication, news, scientific video, and EU legal text. Four evaluation protocols are applied, and factuality is verified using four complementary methods, namely \textcolor{red}{\uline{FactCheck}}\textsuperscript{(\textit{e})}, SAFE, FActScore, and VeriScore, each decomposing summaries into atomic claims and verifying them against retrieved evidence. Human summaries consistently score higher on informativeness and faithfulness. LLMs perform comparably on coherence and conciseness, suggesting that fluency gains do not directly translate to content gains. The self-preference analysis shows that \textcolor{red}{\uline{retaining self-judgments primarily inflates model scores on informativeness and faithfulness rather than on coherence or conciseness,}}\textsuperscript{(\textit{f})} narrowing the gap between human and model outputs on those dimensions. Linguistic analysis confirms lower lexical diversity and less varied discourse structures in LLM outputs.
\end{tcolorbox}
\caption{Gemini-generated summary. Errors (\textit{e}) and (\textit{f}) are marked in \textcolor{red}{\uline{red with underline}}.}
\label{fig:case_gemini}
\end{figure*}

\begin{figure*}[t]
\centering
\begin{tcolorbox}[width=\linewidth,
  colback=violet!4!white,
  colframe=violet!40!black,
  coltitle=black,
  colbacktitle=violet!15!white,
  fonttitle=\bfseries,
  title=Qwen-Generated Summary,
]
This paper revisits the question of whether LLMs have effectively solved summarization through a four-track evaluation spanning five datasets and five frontier models. The evaluation framework covers human annotation, LLM judging, factuality verification, and linguistic analysis. Human summaries consistently outperform model outputs on informativeness and faithfulness. LLMs match or exceed human references on surface-level coherence and conciseness. Factuality analysis reveals that \textcolor{red}{\uline{the gap between human and model summaries is most pronounced on SAFE (0.08 to 0.13), which verifies claims through iterative retrieval against web-sourced evidence.}}\textsuperscript{(\textit{g})} The authors argue that while LLMs raise the floor of summarization quality, the ceiling defined by information density, factual reliability, and linguistic richness remains out of reach.
\end{tcolorbox}
\caption{Qwen-generated summary. Error (\textit{g}) is marked in \textcolor{red}{\uline{red with underline}}.}
\label{fig:case_qwen}
\end{figure*}

\begin{figure*}[t]
\centering
\begin{tcolorbox}[width=\linewidth,
  colback=teal!4!white,
  colframe=teal!40!black,
  coltitle=black,
  colbacktitle=teal!15!white,
  fonttitle=\bfseries,
  title=Kimi-Generated Summary,
]
This position paper presents evidence against the claim that LLMs have solved summarization, drawing on a multi-track evaluation that covers five datasets and five state-of-the-art models. The study applies human annotation, LLM-as-Judge scoring, factuality verification using four complementary methods, and linguistic analysis at lexical, syntactic, and discourse levels. Human summaries retain clear advantages in informativeness and faithfulness, and LLMs score comparably on coherence and conciseness. Factuality analysis shows that LLMs are more prone to errors on reasoning-heavy claims. Linguistic analysis documents stylistic homogeneity in LLM outputs, finding that \textcolor{red}{\uline{this homogeneity is most pronounced between GPT and Claude, which display near-identical lexical diversity and syntactic depth profiles across all five evaluated datasets.}}\textsuperscript{(\textit{h})} Discourse-level analysis further shows that model summaries follow source document order more closely and compress less aggressively than human references.
\end{tcolorbox}
\caption{Kimi-generated summary. Error (\textit{h}) is marked in \textcolor{red}{\uline{red with underline}}.}
\label{fig:case_kimi}
\end{figure*}

\begin{table*}[t]
\centering
\small
\setlength{\tabcolsep}{20pt}
\resizebox{\textwidth}{!}{%
\begin{tabular}{clp{4.2cm}lp{4.2cm}}
\toprule
Label & Model & Hallucinated text (abbreviated) & Error type & Correct information \\
\midrule
(\textit{a}) & GPT    & ``Claude shows the smallest gap \ldots{} followed by GPT''
    & Finding inversion
    & The paper reports GPT as showing the smallest gap relative to human references in human evaluation \\[4pt]
(\textit{b}) & GPT    & ``controls for verbosity bias and self-preference bias''
    & Terminology substitution
    & The protocol mitigates position bias (via presentation order randomization), not verbosity bias \\[4pt]
(\textit{c}) & Claude & ``inter-annotator agreement monitored via Cohen's kappa''
    & Metric misattribution
    & The paper uses Krippendorff's $\alpha$ ($\geq 0.70$), not Cohen's kappa \\[4pt]
(\textit{d}) & Claude & ``eight metrics organized at three levels''
    & Numerical error
    & The paper reports seven linguistic metrics: TTR, MATTR, tree depth, NP modifiers, topic progression, information ordering, and compression ratio \\[4pt]
(\textit{e}) & Gemini & ``FactCheck''
    & Name substitution
    & The factuality method is FaStFact, not FactCheck; both are factuality verification tools but differ in methodology \\[4pt]
(\textit{f}) & Gemini & ``inflates model scores on informativeness and faithfulness''
    & Finding inversion
    & Self-preference bias inflates scores on coherence and conciseness (form-oriented dimensions) \\[4pt]
(\textit{g}) & Qwen   & ``most pronounced on SAFE (0.08 to 0.13)''
    & Fabricated quantification
    & The paper reports aggregate margins of 0.04 to 0.13 across the four methods and does not attribute any sub-range to a specific method; both ``SAFE'' and ``0.08 to 0.13'' are unsupported by the paper text \\[4pt]
(\textit{h}) & Kimi   & ``most pronounced between GPT and Claude \ldots{} near-identical profiles''
    & Unsupported specificity
    & The paper attributes stylistic homogeneity to LLM families in general; no specific model pair is identified as most similar \\
\bottomrule
\end{tabular}
}
\caption{Errors identified in the five model-generated summaries shown in \autoref{fig:case_gpt} through \autoref{fig:case_kimi}. Letter labels correspond to the annotated spans in the summary boxes.}
\label{tab:case_study_errors}
\end{table*}

\section{Prompt Templates and Annotation Guidelines}
\label{appendix:prompts}

This section collects all prompt templates and annotation guidelines used in our evaluation pipeline. \autoref{fig:prompt_cnnsum} through \autoref{fig:prompt_eurlex} present the per-dataset summary-generation templates (all models receive the same prompt for a given dataset, with the source document inserted at the placeholder). \autoref{fig:human_eval_guidelines} provides the annotation guidelines for human evaluators, and \autoref{fig:judge_prompt} shows the LLM-as-Judge prompt.

\begin{figure*}[t]
\centering
\begin{tcolorbox}[width=\linewidth,
  colback=gray!5!white,
  colframe=gray!15!black,
  coltitle=black,
  colbacktitle=gray!25!white,
  fonttitle=\bfseries,
  title=CNNSum Summarization Prompt (Chinese),
]
\begin{CJK}{UTF8}{gbsn}
你是一位专业的中文长篇小说摘要撰写者，具备跨章节叙事追踪与多线情节梳理能力。你的任务是根据提供的完整小说文本，撰写一段结构清晰、内容完整、客观中立的故事梗概，使读者无需阅读原文即可全面把握全篇内容。\\
\textbf{输入：} \\
一部长篇中文网络小说的完整正文，篇幅可能超过十万字。 \\
\textbf{要求：} \\
1. \textit{核心要素覆盖。} 梗概须涵盖以下要素：(a) 主角的身份、出身背景与核心目标或动机；(b) 与情节直接相关的世界观设定（如特殊力量体系、社会结构、地理环境）；(c) 贯穿全篇的主线情节及其阶段性发展脉络；(d) 不少于两个关键转折点或高潮事件，并简要说明其对人物处境与故事走向的影响；(e) 主要配角的身份、立场及其与主角的关系演变；(f) 故事的最终结局或走向（若原文已完结）。 \\
2. \textit{叙事顺序与结构。} 按故事内时间顺序叙述，确保因果链条清晰；多线叙事时以主线为骨架串联，支线信息在与主线交汇处简要带过，避免线索断裂。 \\
3. \textit{客观性约束。} 严格不得加入原文未出现的情节、人物、地点、时间或因果关系，不得添加任何主观评价、文学赏析、情感感叹或对结局的推测，不得使用任何评论性词汇。 \\
4. \textit{语言规范。} 使用规范的中文书面语，避免口语化、网络流行语与作者点评式语气。人名、地名、特殊术语及专有名词须与原文用字完全一致，不得替换为常见同义词。 \\
\textbf{小说正文：} \texttt{\{document\}}
\end{CJK}
\end{tcolorbox}
\caption{Summary-generation prompt for CNNSum (Chinese novel summarization).}
\label{fig:prompt_cnnsum}
\end{figure*}

\begin{figure*}[t]
\centering
\begin{tcolorbox}[width=\linewidth,
  colback=gray!5!white,
  colframe=gray!15!black,
  coltitle=black,
  colbacktitle=gray!25!white,
  fonttitle=\bfseries,
  title=SciNews Summarization Prompt (English),
]
You are an expert science communicator with experience translating peer-reviewed research for general audiences. Your task is to write a lay summary of the following scientific research paper for a reader with no specialized background. \\
\textbf{Inputs:} \\
The full text of a peer-reviewed research paper. \\
\textbf{Instructions:} \\
1. \textit{Structure.} Organize the summary around four elements in sequence: (a) the scientific question or problem and why it matters in everyday terms; (b) the approach or method, described in accessible language with a concrete analogy where useful; (c) the main findings, with key quantitative results paraphrased in plain language while preserving their precision; and (d) the broader significance, including any caveats, limitations, or open questions that the authors explicitly acknowledge. \\
2. \textit{Accessibility.} Replace specialized terminology with plain-language equivalents wherever possible. For domain-specific terms that cannot be avoided (e.g., names of specific molecules, instruments, species, or statistical procedures), provide a brief parenthetical definition on first use. \\
3. \textit{Hedging fidelity.} Preserve the speculative register of hedged statements (e.g., ``the authors suggest'' rather than ``the study proves''; ``may be associated with'' rather than ``causes''). Do not strengthen claims beyond what the paper states. \\
4. \textit{No sensationalism.} Avoid framings such as ``breakthrough,'' ``revolutionary,'' or ``solves'' unless the authors themselves use such language. \\
5. \textit{Numerical fidelity.} Preserve reported values, units, and statistical qualifiers (e.g., confidence intervals, $p$-values, sample sizes). Do not paraphrase quantities in a way that changes their precision or implied uncertainty. \\
6. \textit{Style.} Write in clear, grammatically correct English prose. \\
\textbf{Paper:} \texttt{\{document\}}
\end{tcolorbox}
\caption{Summary-generation prompt for SciNews (lay science summarization).}
\label{fig:prompt_scinews}
\end{figure*}

\begin{figure*}[t]
\centering
\begin{tcolorbox}[width=\linewidth,
  colback=gray!5!white,
  colframe=gray!15!black,
  coltitle=black,
  colbacktitle=gray!25!white,
  fonttitle=\bfseries,
  title=DiverseSumm Summarization Prompt (English),
]
You are an experienced news editor with expertise in cross-source synthesis. Your task is to write a single coherent summary that integrates information from the following set of news articles covering the same event or topic, capturing both points of consensus and the diverse perspectives that distinguish individual sources. \\
\textbf{Inputs:} \\
A set of news articles reporting on the same event or topic. \\
\textbf{Instructions:} \\
1. \textit{Core facts.} Identify the facts shared across articles (who, what, when, where) and use them as the backbone of the summary. \\
2. \textit{Diverse perspectives.} Beyond the consensus, surface information that appears in only a subset of the articles, including (a) unique details, quotations, or background context contributed by individual sources; (b) different framings or interpretations of the same event; and (c) regional, partisan, or institutional angles that distinguish how outlets cover the story. \\
3. \textit{Conflicts and discrepancies.} When sources report conflicting details on a materially significant fact (e.g., differing casualty numbers, opposing characterizations of an actor's motives), present both versions with brief attribution rather than silently adopting the most widely repeated one. \\
4. \textit{Non-redundancy.} Report each shared fact once and in its most complete form; do not repeat information solely because it appears in multiple sources. \\
5. \textit{Editorial neutrality.} Do not introduce information, editorial judgments, or conclusions absent from all provided articles. Attribute any contested or interpretive claim to its source. \\
6. \textit{Style.} Write in clear, neutral, third-person journalistic English. \\
\textbf{Articles:} \texttt{\{documents\}}
\end{tcolorbox}
\caption{Summary-generation prompt for DiverseSumm (multi-document news summarization).}
\label{fig:prompt_diversesumm}
\end{figure*}

\begin{figure*}[t]
\centering
\begin{tcolorbox}[width=\linewidth,
  colback=gray!5!white,
  colframe=gray!15!black,
  coltitle=black,
  colbacktitle=gray!25!white,
  fonttitle=\bfseries,
  title=VISTA Summarization Prompt (English),
]
You are an expert science writer familiar with the conventions of academic abstracts and the structure of AI/ML conference presentations (e.g., NeurIPS, ICML, ACL, CVPR). Your task is to write an abstract-style summary of the scientific content in the following recorded conference talk, drawing on both the video and its transcript. \\
\textbf{Inputs:} \\
1. A recorded scientific conference presentation video, including the speaker's slides and any on-screen demonstrations. \\
2. A verbatim or near-verbatim transcript of the same presentation. \\
\textbf{Instructions:} \\
1. \textit{Abstract structure.} Organize the summary around five elements in order: (a) the scientific motivation, framed as an existing limitation or open question; (b) the proposed approach, model, or experimental design, including any architectural or methodological innovations; (c) the evaluation setup, with datasets, baselines, and metrics where mentioned; (d) the main quantitative or qualitative results; and (e) the primary conclusion or claimed significance. \\
2. \textit{Multimodal integration.} Use the video to identify visual content displayed during the talk (slides, figures, tables, diagrams, demos) and integrate it with the transcript. When a visual element complements a verbal description (e.g., a figure illustrates a result, an architecture slide accompanies a method explanation), draw on both sources. \\
3. \textit{Technical precision.} Preserve technical terminology, model names, dataset names, and metric names exactly as they appear in the video or transcript, together with their reported numerical values and units (e.g., accuracy points, F1, BLEU, perplexity). \\
4. \textit{Faithfulness.} Do not report observations or conclusions absent from both the video and the transcript. Do not extrapolate from a specific result to a broader claim that the speaker does not make. \\
5. \textit{Style.} Write in formal, concise academic English in the present tense. \\
\textbf{Video:} \texttt{\{video\}} \\
\textbf{Transcript:} \texttt{\{transcript\}}
\end{tcolorbox}
\caption{Summary-generation prompt for VISTA (video-to-text scientific summarization).}
\label{fig:prompt_vista}
\end{figure*}

\begin{figure*}[t]
\centering
\begin{tcolorbox}[width=\linewidth,
  colback=gray!5!white,
  colframe=gray!15!black,
  coltitle=black,
  colbacktitle=gray!25!white,
  fonttitle=\bfseries,
  title=EurLexSum Summarization Prompt (Multilingual),
]
You are an expert legal summarizer with training in EU administrative and legislative documents, comparable to staff producing the official summaries of legislation published on the EUR-Lex portal. Your task is to write a concise, accurate summary of the following EU legislative or regulatory document. The summary must be written in the same language as the source document. \\
\textbf{Inputs:} \\
The full text of an EU legislative or regulatory document in \texttt{\{language\}}. \\
\textbf{Instructions:} \\
1. \textit{Required elements.} The summary must address the following in a manner appropriate to the instrument type: (a) the type and title of the instrument (e.g., Regulation, Directive, Decision, Recommendation, Opinion) together with its legal basis in the Treaties; (b) the subject matter and the territorial, sectoral, or personal scope of application; (c) the primary obligations, prohibitions, or entitlements established, with reference to the relevant articles where appropriate; (d) the key defined terms that govern interpretation of the instrument; and (e) the entry-into-force date and any transitional or sunset provisions affecting implementation timing. \\
2. \textit{Recitals vs.\ operative articles.} Treat the operative articles as the primary source of binding content. Recitals may be drawn on to clarify legislative intent or context, but conclusions stated only in recitals must not be presented as legal obligations. \\
3. \textit{Legal precision.} Do not paraphrase defined terms in a way that alters their regulatory meaning. Preserve the distinction between obligations (\textit{shall}, \textit{must}), permissions (\textit{may}), and recommendations (\textit{should}) as used in the original text. \\
4. \textit{No extrapolation.} Do not introduce information, legal interpretations, cross-references, or comparisons with other instruments that are absent from the source document. \\
5. \textit{Naming conventions.} Institutional names (e.g., the Council, the Commission, the Parliament, the Court of Justice) and proper nouns must appear in the same form as in the original. Numerical identifiers of articles, regulations, and directives must be preserved exactly. \\
6. \textit{Style.} Write in formal legal prose in \texttt{\{language\}}.  \\
\textbf{Legislative document:} \texttt{\{document\}}
\end{tcolorbox}
\caption{Summary-generation prompt for EurLexSum (multilingual legal summarization). The \texttt{\{language\}} placeholder is replaced with the target language name (e.g., English, German, or French).}
\label{fig:prompt_eurlex}
\end{figure*}

\begin{figure*}[t]
\centering
\begin{tcolorbox}[width=\linewidth,
  colback=gray!5!white,
  colframe=gray!15!black,
  coltitle=black,
  colbacktitle=gray!25!white,
  fonttitle=\bfseries,
  fontupper=\small,
  title=Human Evaluation Annotation Guidelines,
]
\textbf{Task overview.} You will be shown a source document and six candidate summaries (labeled Summary 1 through Summary 6) in randomized order. System identities are hidden, and the presentation order varies across samples. Evaluate each summary on four dimensions independently, then provide an overall ranking. \\[4pt]
\textbf{Dimension definitions and scoring anchors.} \\
1. \textbf{Informativeness.} Does the summary capture the key information from the source? Key information refers to the points that a knowledgeable reader of the source would consider essential.
\begin{itemize}[leftmargin=15pt,itemsep=0pt,topsep=2pt,parsep=0pt]
\item \textit{Score 1:} Misses most key points or covers only a small fraction of the main content.
\item \textit{Score 3:} Covers the most important points but omits several secondary details relevant to a complete understanding.
\item \textit{Score 5:} Captures all major points and key supporting details without omitting information that would change the reader's understanding.
\end{itemize}
2. \textbf{Faithfulness.} Is every claim in the summary factually accurate? Verify each claim first against the source document; for claims about widely accepted general knowledge that the source presupposes, supplement with credible external references.
\begin{itemize}[leftmargin=15pt,itemsep=0pt,topsep=2pt,parsep=0pt]
\item \textit{Score 1:} Contains multiple claims that are factually incorrect or that cannot be verified from the source or credible external references.
\item \textit{Score 3:} Mostly accurate but contains one or two statements that are imprecise or weakly supported.
\item \textit{Score 5:} Every claim is factually accurate and supported by the source or verifiable through credible external references.
\end{itemize}
3. \textbf{Coherence.} Is the summary well-organized, grammatically correct, and easy to read? Coherence concerns surface readability and linguistic quality, not length or comprehensiveness; a longer summary is not less coherent for being longer.
\begin{itemize}[leftmargin=15pt,itemsep=0pt,topsep=2pt,parsep=0pt]
\item \textit{Score 1:} Difficult to follow due to grammatical errors, disorganized structure, or unclear references.
\item \textit{Score 3:} Generally readable but contains some awkward transitions or minor grammatical issues.
\item \textit{Score 5:} Well-structured, grammatically correct, and reads naturally throughout.
\end{itemize}
4. \textbf{Conciseness.} Does the summary avoid unnecessary repetition and verbosity? Conciseness concerns brevity and absence of redundancy, not grammatical quality; a grammatically rough but tight summary should not lose conciseness points for grammar alone.
\begin{itemize}[leftmargin=15pt,itemsep=0pt,topsep=2pt,parsep=0pt]
\item \textit{Score 1:} Highly repetitive or padded with irrelevant content.
\item \textit{Score 3:} Mostly concise but contains some redundant phrases or minor verbosity.
\item \textit{Score 5:} Conveys all essential information without repetition, padding, or off-topic content.
\end{itemize}
\vspace{4pt}
\textbf{Evaluation procedure.} \\
1. \textit{Per-dimension Likert scores.} For each dimension, assign each summary a score from 1 (very poor) to 5 (excellent). Use intermediate scores (2 and 4) for summaries that fall between the described anchors. \\
2. \textit{Overall ranking.} After completing all dimension scores, rank all six summaries from 1 (worst) to 6 (best) based on your judgment of overall quality. Ties are not permitted. \\ \\
\textbf{Important reminders.} \\
1. \textit{Dimension independence.} Evaluate each dimension independently; do not let your judgment on one dimension influence scores on another. \\
2. \textit{Summary independence.} Score each summary on its own merits; do not let your impression of an earlier summary anchor your scores for later ones, and revisit earlier scores if a later comparison reveals a calibration drift. \\
3. \textit{Faithfulness sources.} Verify summary claims first against the source document. Where the summary makes a claim about widely accepted general knowledge, supplement with credible external references (e.g., Wikipedia, peer-reviewed papers, reference works). \\
4. \textit{Edge cases.} For truncated summaries, score the content present and reflect any disruption to closure under coherence. For summaries with extraneous formatting (bullet points, headings, model markers), penalize only when the formatting disrupts reading. \\
5. \textit{Cross-lingual evaluation.} If the source is in a language other than English, evaluate in that language and apply the same criteria. \\
6. \textit{No AI assistance.} Do not use any AI tool for evaluation. \\
\end{tcolorbox}
\caption{Human-evaluation guidelines. Each annotator receives these instructions together with a source document and six candidate summaries presented in blind, randomized order.}
\label{fig:human_eval_guidelines}
\end{figure*}

\begin{figure*}[t]
\centering
\begin{tcolorbox}[width=\linewidth,
  colback=gray!5!white,
  colframe=gray!15!black,
  coltitle=black,
  colbacktitle=gray!25!white,
  fonttitle=\bfseries,
  fontupper=\small,
  title=LLM-as-Judge Evaluation Prompt,
]
You are an expert summary evaluator. You will be given a source document and six candidate summaries (labeled Summary A through Summary F) in randomized order. Your task is to score all six summaries on each of four dimensions independently, then provide an overall ranking of summary quality. \\[4pt]
\textbf{Dimension definitions and scoring anchors.} \\
1. \textbf{Informativeness.} Does the summary capture the key information from the source? Key information refers to the points that a knowledgeable reader of the source would consider essential.
\begin{itemize}[leftmargin=15pt,itemsep=0pt,topsep=2pt,parsep=0pt]
\item \textit{Score 1:} Misses most key points or covers only a small fraction of the main content.
\item \textit{Score 3:} Covers the most important points but omits several secondary details relevant to a complete understanding.
\item \textit{Score 5:} Captures all major points and key supporting details without omitting information that would change the reader's understanding.
\end{itemize}
2. \textbf{Faithfulness.} Is every claim in the summary factually accurate? Verify each claim first against the source document; for claims about widely accepted general knowledge that the source presupposes, supplement with credible external references.
\begin{itemize}[leftmargin=15pt,itemsep=0pt,topsep=2pt,parsep=0pt]
\item \textit{Score 1:} Contains multiple claims that are factually incorrect or that cannot be verified from the source or credible external references.
\item \textit{Score 3:} Mostly accurate but contains one or two statements that are imprecise or weakly supported.
\item \textit{Score 5:} Every claim is factually accurate and supported by the source or verifiable through credible external references.
\end{itemize}
3. \textbf{Coherence.} Is the summary well-organized, grammatically correct, and easy to read? Coherence concerns surface readability and linguistic quality, not length or comprehensiveness; a longer summary is not less coherent for being longer.
\begin{itemize}[leftmargin=15pt,itemsep=0pt,topsep=2pt,parsep=0pt]
\item \textit{Score 1:} Difficult to follow due to grammatical errors, disorganized structure, or unclear references.
\item \textit{Score 3:} Generally readable but contains some awkward transitions or minor grammatical issues.
\item \textit{Score 5:} Well-structured, grammatically correct, and reads naturally throughout.
\end{itemize}
4. \textbf{Conciseness.} Does the summary avoid unnecessary repetition and verbosity? Conciseness concerns brevity and absence of redundancy, not grammatical quality; a grammatically rough but tight summary should not lose conciseness points for grammar alone.
\begin{itemize}[leftmargin=15pt,itemsep=0pt,topsep=2pt,parsep=0pt]
\item \textit{Score 1:} Highly repetitive or padded with irrelevant content.
\item \textit{Score 3:} Mostly concise but contains some redundant phrases or minor verbosity.
\item \textit{Score 5:} Conveys all essential information without repetition, padding, or off-topic content.
\end{itemize} 
\textbf{Instructions.} \\
1. \textit{Read the source.} Read the source document carefully before evaluating any summary. \\
2. \textit{Per-dimension scoring.} For each of the four dimensions, assign each of the six summaries a Likert score from 1 (very poor) to 5 (excellent). Multiple summaries may receive the same score on a given dimension. Use intermediate scores (2 and 4) for summaries that fall between the described anchors. \\
3. \textit{Overall ranking.} After scoring all four dimensions, provide an overall ranking of all six summaries from worst (rank 1) to best (rank 6). Each rank must be used exactly once; ties are not permitted. \\
4. \textit{Bias mitigation.} Score each summary on its own content, not its position in the list (A through F) or its surface style. Do not let stylistic familiarity favor one summary over another. Do not let your impression of an earlier summary anchor your scores for later ones. \\
5. \textit{Dimension independence.} Evaluate each dimension independently. Do not allow your assessment of one dimension to influence your scores on another. \\ [4pt]
\textbf{Output format.} \\
Return your scores and ranking in the following exact format: \\
\texttt{Informativeness: A=[1-5], B=[1-5], C=[1-5], D=[1-5], E=[1-5], F=[1-5]} \\
\texttt{Faithfulness: A=[1-5], B=[1-5], C=[1-5], D=[1-5], E=[1-5], F=[1-5]} \\
\texttt{Coherence: A=[1-5], B=[1-5], C=[1-5], D=[1-5], E=[1-5], F=[1-5]} \\
\texttt{Conciseness: A=[1-5], B=[1-5], C=[1-5], D=[1-5], E=[1-5], F=[1-5]} \\
\texttt{Overall: [A/B/C/D/E/F] < [A/B/C/D/E/F] < ... < [A/B/C/D/E/F]} \\[4pt]
\textbf{Source document:} \texttt{\{document\}} \\[2pt]
\textbf{Summary A:} \texttt{\{summary\_a\}} \\
\textbf{Summary B:} \texttt{\{summary\_b\}} \\
\textbf{Summary C:} \texttt{\{summary\_c\}} \\
\textbf{Summary D:} \texttt{\{summary\_d\}} \\
\textbf{Summary E:} \texttt{\{summary\_e\}} \\
\textbf{Summary F:} \texttt{\{summary\_f\}}
\end{tcolorbox}
\caption{LLM-as-Judge evaluation prompt. The six candidate summaries are independently randomized for each sample and judge.}
\label{fig:judge_prompt}
\end{figure*}
\end{document}